\documentclass[sigplan,screen]{acmart}

\copyrightyear{2025}
\acmYear{2025}
\setcopyright{cc}
\setcctype{by}
\acmConference[ASPLOS '25]{Proceedings of the 30th ACM International Conference on Architectural Support for Programming Languages and Operating Systems, Volume 3}{March 30-April 3, 2025}{Rotterdam, Netherlands}
\acmBooktitle{Proceedings of the 30th ACM International Conference on Architectural Support for Programming Languages and Operating Systems, Volume 3 (ASPLOS '25), March 30-April 3, 2025, Rotterdam, Netherlands}
\acmDOI{10.1145/3676642.3736118}
\acmISBN{979-8-4007-1080-3/2025/03}

\usepackage[normalem]{ulem}
\usepackage{makecell}
\usepackage{booktabs}
\usepackage{multirow}
\usepackage{amsmath, amsfonts, amsthm}
\usepackage{textcomp}
\usepackage{textgreek}
\usepackage{enumitem}
\setlist{topsep=0pt}
\usepackage{verbatim}
\usepackage{xspace}
\usepackage{listings}
\usepackage{xcolor}
\makeatletter 
\@namedef{ver@lineno.sty}{9999/12/31}
\@namedef{opt@lineno.sty}{}
\makeatother
\usepackage[frozencache]{minted}
\usepackage{mathtools}
\usepackage[per-mode=symbol, range-units=single, mode=text]{siunitx}

\usepackage[noabbrev, capitalise, nameinlink]{cleveref}
\crefname{listing}{Listing}{Listings}
\crefname{listing}{Listing}{Listings}

\usepackage{stackrel}

\usepackage{algorithm}
\usepackage{algpseudocodex}

\newcommand{\algorithmiccontinue}{\textbf{continue}}

\usepackage{flushend}

\newcommand{\sys}{Syno\xspace}
\newcommand{\tablefontsize}{\small}
\newcommand{\prim}[1]{\textsc{#1}}

\newlength{\arrow}
\newcommand*{\alignedleftarrow}[1]{\mathmakebox[\arrow]{\stackrel{#1}{\longleftarrow}}}

\title{\sys: Structured Synthesis for Neural Operators}

\author{Yongqi Zhuo}
\orcid{0009-0004-2229-2750}
\affiliation{
  \institution{Tsinghua University}
  \city{Beijing}
  \country{China}
}
\email{zhuoyq21@mails.tsinghua.edu.cn}
\authornote{Both authors contributed equally to the paper.}

\author{Zhengyuan Su}
\orcid{0009-0003-0121-2128}
\affiliation{
  \institution{Tsinghua University}
  \city{Beijing}
  \country{China}
}
\email{su-zy21@mails.tsinghua.edu.cn}
\authornotemark[1]

\author{Chenggang Zhao}
\orcid{0009-0005-2297-0790}
\affiliation{
  \institution{Tsinghua University}
  \city{Beijing}
  \country{China}
}
\email{zhaocg21@mails.tsinghua.edu.cn}

\author{Mingyu Gao}
\orcid{0000-0001-8433-7281}
\affiliation{%
  \institution{Tsinghua University}
  \city{Beijing}
  \country{China}
}
\affiliation{%
  \institution{Shanghai Artificial Intelligence Lab}
  \city{Shanghai}
  \country{China}
}
\affiliation{%
  \institution{Shanghai Qi Zhi Institute}
  \city{Shanghai}
  \country{China}
}
\email{gaomy@tsinghua.edu.cn}

\begin{document}

\begin{abstract}
The desires for better prediction accuracy and higher execution performance in neural networks never end. 
Neural architecture search (NAS) and tensor compilers are two popular techniques to optimize these two goals, but they are both limited to composing or optimizing existing manually designed operators rather than coming up with completely new designs. 
In this work, we explore the less studied direction of neural operator synthesis, which aims to automatically and efficiently discover novel neural operators with better accuracy and/or speed. 
We develop an end-to-end framework \sys, to realize practical neural operator synthesis.
\sys makes use of a novel set of fine-grained primitives defined on tensor dimensions, which ensure various desired properties to ease model training, and also enable expression canonicalization techniques to avoid redundant candidates during search. 
\sys further adopts a novel guided synthesis flow to obtain valid operators matched with the specified input/output dimension sizes, and leverages efficient stochastic tree search algorithms to quickly explore the design space. 
We demonstrate that \sys discovers better operators with average speedups of $1.37\times$ to $2.06\times$ on various hardware and compiler choices, while keeping less than 1\% accuracy loss even on NAS-optimized models. 

\end{abstract}



\maketitle 

\section{Introduction}

Deep learning with neural networks (NNs) has been a surprisingly effective algorithm breakthrough to handle many challenging tasks in various domains. Since its emergence in the last decade, people have been continuously seeking to improve both the quality (in terms of, e.g., prediction accuracy) and the performance (in terms of training and inference time) of NN models, in order to adapt them to more complicated real-world scenarios with lower computational cost. 

Two complementary research paradigms have been developed. 
To systematically design new NN models with better accuracy quality, neural architecture search (NAS)~\cite{nas-reinforcement, nas-learning, nas-progressive, nas-survey} uses deep learning algorithms themselves to automatically discover promising model structures~\cite{mnasnet, efficientnet, fbnet, primer}. Given a backbone network topology, NAS explores how to construct the basic cells in the model using combinations of basic operators like convolutions and pooling. The optimization goal is either pure accuracy, or a balance between accuracy and speed~\cite{fbnet, proxyless, hardware-aware, mnasnet}.
In contrast, to improve training and inference speeds, tensor compilers~\cite{tvm, pet, ragan2013halide, zheng2020ansor, akg} aim to optimize the implementation of low-level loop nests of each operator in an NN model. Various general and specialized compile-time optimizations are applied to the operator, without altering its functional semantics. 

We notice that a new direction orthogonal to the above two has not been fully explored, namely to \emph{synthesize novel neural operators at a fine granularity}, with the goal to automatically and efficiently discover new operators beyond existing standard types (e.g., convolutions), to improve accuracy quality and/or execution performance. 
NAS only composes its cell structures using \emph{existing} operators, and tensor compilers only explore \emph{semantically equivalent} variants of the original operators.  
We envision that these three complementary approaches could be used together. For example, starting from a NAS-discovered network topology, we synthesize novel operators to replace the original ones in the model, and finally leverage tensor compilers to optimize their execution speeds on specific hardware backends.

Neural operator synthesis can be viewed as a domain-specific form of program synthesis~\cite{gulwani2017program}, a classic topic in computer science. 
Generic program synthesis approaches face the issue of scalability and do not easily allow different target semantics from the specification. Early attempts of program synthesis for NNs~\cite{jin2022neural, ma2022searching} are still limited to composing with coarse-grained basic operators and leave large potentials unexploited.
More specifically, to explore a sufficiently large design space, \emph{fine-grained synthesis} that composes directly from the very basic programming language atoms is desired. This is highly challenging. 
First, the arbitrarily composed candidates would be very unlikely to satisfy the common properties of neural operators, such as differentiability, no replication or discard of tensor elements, etc. 
Second, there will be enormous redundant synthesized operator candidates with the same or similar semantics. Such equivalent semantics are already efficiently explored by tensor compilers, so we want to prune them out and focus on discovering new operators. 
Finally, program synthesis has complicated search spaces, and the search for neural operators has the specialized goals of better inference accuracy and performance, which are different from the strict and clear correctness requirement in traditional synthesis. A new search method is thus needed.

We develop \emph{an end-to-end, automatic, and efficient neural operator synthesis framework}, \sys. 
It takes a given backbone NN topology, and searches for novel linear operators to replace the original ones in the model, in order to improve prediction accuracy and/or execution performance. 
\sys addresses the aforementioned challenges with several key techniques. 
First, it makes use of a novel set of \emph{fine-grained primitives} to synthesize new operators. The primitive semantics are defined on tensor coordinates (i.e., dimensions). They maintain tensor semantics and exhibit high-quality properties for neural operators, while not sacrificing expressiveness. 
Second, \sys leverages \emph{expression simplification and canonicalization} techniques to analyze and eliminate most of the redundancies when synthesizing operators, especially to avoid redoing tensor compiler optimizations. 
Finally, the design space search process is made more structured, which iteratively samples and adds new primitives to compose operator candidates. We formulate it as a Markov decision process and leverage the efficient \emph{Monte Carlo Tree Search} algorithm~\cite{coulom2007mcts, gaudel2010fuse, hsu2020mcts}. 
We further propose a novel metric of \emph{shape distance} to guide the synthesis towards matching with the required input/output tensor shapes, so that the synthesized operator candidate is valid to be used in the backbone model. 
Two code generators targeting PyTorch~\cite{ansel2024pytorch} and TVM~\cite{tvm} are built for accuracy and speed evaluation of \sys-discovered operators.

Evaluated on five vision models and GPT-2, \sys discovers faster operators than standard convolutions and matrix multiplications, even on NAS-optimized backbone models.  
Within 1\% accuracy loss on CIFAR-100, using \sys-optimized operators exhibits $2.06\times$, $1.72\times$, and $1.47\times$ speedups on average on mobile CPUs, mobile GPUs, and server GPUs, respectively, with the TVM backend, compared to the original models.
With \texttt{torch.compile}~\cite{ansel2024pytorch}, the speedups are $1.37\times$, $1.62\times$, and $1.60\times$.
On ImageNet, \sys-optimized operators achieve up to $4.73\times$ and $1.94\times$ speedups when compiled with TVM and \texttt{torch.compile}, respectively, with 1\% to 2\% accuracy loss.
\sys also accelerates GPT-2 training by $1.1\times$ and improves the language perplexity metric from 111 to 99. 
We also investigate the discovered operators and find novel and efficient semantics with interesting neural algorithm insights.

\section{Background and Related Work}
\label{sec:background}

In this section, we briefly introduce the three most related concepts: neural architecture search, tensor compilers, and program synthesis, all of which can be used to better design and implement neural network operators.

\subsection{Neural Architecture Search}
\label{sec:background:nas}

As neural networks (NNs) are being applied to more and more domains, the needs of designing specific NN models are becoming increasingly prevalent.
\emph{Neural architecture search} (NAS) has emerged consequently to automatically design new model structures~\cite{nas-reinforcement, nas-learning, nas-progressive, nas-survey}, and indeed, many of the recently proposed models that showed state-of-the-art accuracy levels were discovered by NAS rather than manually crafted~\cite{mnasnet, efficientnet, fbnet, primer}. 
NAS typically defines a highly modular search space by dividing the backbone model topology into basic units (called \emph{cells}) of various sizes. It then proceeds to explore how to construct each cell by composing from several types of basic layers (a.k.a., \emph{operators}) like convolutions, matrix multiplications (matmuls), and pooling. Throughout the search, the accuracy levels of the candidate cell structures are continuously evaluated. 
With such an automated flow, NAS is able to efficiently explore a large design space, and hence discover NN architectures with potentially higher accuracy than manually designed models.

Besides solely focusing on accuracy, \emph{performance-aware NAS} methods aim to strike a better balance between prediction accuracy and execution speed~\cite{fbnet, proxyless, hardware-aware, mnasnet}. 
Specifically, they inherit the design space from traditional NAS while integrating hardware efficiency metrics. By considering factors such as latency alongside accuracy, the search process could yield model architectures that are not only high-quality but also high-performance\footnote{Throughout this paper, we use ``quality'' for model accuracy, and ``performance'' for execution speed.} on particular hardware.

We emphasize that both traditional and performance-aware NAS methods only \emph{compose existing operators}, such as convolutions and matmuls, in a coarse-grained black-box way. Thus they are limited by these computationally expensive operators. 
The lack of flexibility to \emph{invent novel operators} leaves ample opportunities for further optimizations, as we will demonstrate in this work.

\subsection{Tensor Compilers}
\label{sec:background:compiler}

At the system level, an NN model is typically represented as a \emph{tensor program}, in which the input/output and intermediate data are all cast as tensors, and a set of operators are applied to them.
As a result, \emph{tensor compilers} have gained great attention to accelerate NN execution, by applying general and specialized compile-time optimizations to compile the operators into high-performance \emph{kernels}\footnote{We use \emph{kernel} to represent a concrete implementation of an \emph{operator}.}~\cite{tvm, pet, ragan2013halide, zheng2020ansor, akg}.

Typically, kernels are written as loop nests, and each tensor compiler has its own intermediate representation (IR) for describing and optimizing kernels. 
We here take Halide~\cite{ragan2013halide} as an example. 
Many tensor compilers use similar IR designs~\cite{tvm, taichi, freetensor, akg}. Halide provides the separation of algorithm and schedule, where the algorithm is purely functional, and the schedule dictates the concrete loop nest implementation involving tiling, vectorization, reordering, etc.
For example, \cref{list:halide-example} shows how we define a convolution in Halide, which is just simplified loop nests operating on specific \emph{coordinates} (i.e., dimensions) of the tensors. 
With this IR, we can flexibly express different tensor computations.

\begin{listing}[h]
\begin{minted}[fontsize=\footnotesize]{c++}
auto [r_Ci, r_K_H, r_K_W] = RDom(0, C_in, 0, K, 0, K);
out(i_N, i_Co, i_H, i_W) +=
  input(i_N, r_Ci, i_H + r_K_H - K / 2, i_W + r_K_W - K / 2)
  * weight(i_Co, r_Ci, r_K_H, r_K_W);
\end{minted}
\caption{The conv2d operator represented in Halide.}
\label{list:halide-example}
\end{listing}

The separation of algorithm and schedule enables tensor compilers to explore the optimization space that is \emph{semantically equivalent} to the original program, i.e., the purely functional computation description as in \cref{list:halide-example}.
This is in contrast to NAS-like approaches that find \emph{semantically inequivalent} programs with better quality and/or better performance, so the two approaches are \emph{orthogonal}.
A recent work, \citet{turner2021neural}, extended tensor compilers, relaxing the equivalence constraint to apply \emph{inequivalent} transformations on loop nests in tensor programs to realize NAS. However, their approach pre-defined only a few simple inequivalent transformations, such as grouping and bottlenecking the range of a loop, thus only exploring a limited search space still in the scope of traditional operators.

\subsection{Program Synthesis}
\label{sec:background:prog-synth}

Program synthesis is an approach that automatically generates a program that complies with several \emph{specifications}, such as a set of input and output example pairs, or a set of assertions~\cite{gulwani2017program}. 
Theoretically speaking, the general concept of program synthesis can be applied to design new NN operators, but there exist several practical gaps.
Traditional program synthesis only treated \emph{correctness} as the target, such as TF-Coder~\cite{shi2022tf} which synthesizes TensorFlow code to help programmers write correct code. But NN models, which are known to tolerate small errors, do not have a clear notion of correctness, while the goal is to improve inference accuracy and/or execution speed.
Also, existing program synthesis approaches can hardly scale, currently limited to considering a highly constrained program space. The complexity of loop nests in typical NN operators is well beyond their capabilities.
We discuss these challenges in more detail in \cref{sec:motiv}.

$\alpha$NAS~\cite{jin2022neural} relaxed the correctness objective to apply goal-directed program synthesis for NAS. They applied transformations to subgraphs in the model, which could generate new operators beyond traditional NAS. But they are still constrained by traditional operators like convolutions and matmuls, so the potential of intra-operator program synthesis remains unexploited. As a result, their speedups were light, as \cref{sec:eval} will show. 
\citet{ma2022searching}, on the other hand, pre-defined some fine-grained primitives common in traditional demosaicking pipelines to perform NAS; however they did not allow freely exploring new operators, either.

\section{Motivation and Challenges}
\label{sec:motiv}

In this work, we aim to automatically and efficiently synthesize novel NN operators from the very basic atoms in programming languages, in hope of discovering new operators that have both high accuracy quality and high execution performance.
Such automatic \emph{neural operator synthesis} is highly profitable. State-of-the-art models today such as transformers and convolutional networks rely heavily on operators like attention and convolution that are constructed based on human insights. Automatic discovery of such operators can potentially create more promising model architectures.

\textbf{Comparison with existing paradigms.}
Neural operator synthesis has a similar goal to performance-aware NAS, but aims to synthesize tensor programs at a much more fine-grained level rather than directly composing known operators like convolutions and matmuls. 
More specifically, operator synthesis involves writing various loop nests and the tensor expressions in the loop body. 
For example, for the convolution in \cref{list:halide-example}, the loops are implicitly defined by the iterators (\texttt{i\_Co}, \texttt{r\_Ci}, etc.), and the tensor expressions are realized with \emph{coordinate expressions}. Coordinate expressions are key to an operator because they specify how tensor elements are arranged and which are involved in the computation. Here, the simple addition of iterators (\texttt{i\_H + r\_K\_H}) implies convolution, and the repeated uses of the reduction iterator (\texttt{r\_Ci}) in two tensors imply contraction (a.k.a., tensor multiplication). 
The rich semantics of coordinate expressions can be exploited to synthesize novel operators. 
We note that such operator synthesis is impossible under existing NAS. Although it is always possible to lower existing operators to nested loops, it is \emph{not} always possible to do the inverse. If a loop nest cannot be decomposed into several existing operators, it is likely we have discovered a novel operator. 

On the other hand, operator synthesis is also significantly different from tensor compilers. 
Existing tensor compilers mostly preserve semantic equivalence as discussed in \cref{sec:background:compiler}. Thus they are unable to discover \emph{new} operators. Actually, in operator synthesis, we deliberately avoid the exploration of semantically equivalent operators (see \cref{sec:pruning}). 
If we synthesize equivalent operators, we would be very likely to redo existing optimizations in tensor compilers. 
In this sense, \emph{tensor compilers and operator synthesis are orthogonal}. We first synthesize novel operators, and then leverage tensor compilers to optimize their execution performance on the particular hardware. 

We view neural operator synthesis as a specialized form of program synthesis in the NN domain. 
While traditional synthesis methods are limited to simple programs, we need to handle more complex operators with various nested loop structures and coordinate expressions. 
On one hand, the degree of freedom in directly writing loop nests and tensor expressions is huge, leading to an extremely large search space.
On the other hand, as in performance-aware NAS, for each candidate operator, we need to assess both its accuracy level and execution speed, both of which require substantial time. 
To measure the accuracy, we have to use real datasets to train the full NN model for several epochs at least. Several theoretical metrics are proposed to predict the accuracy potential with minimum training cost~\cite{zero-cost-nas, snip, fisherp, econas}, but we find them to perform poorly in reality, especially for irregular operators we aim to construct.
To evaluate the speed, we need to generate an optimized implementation of the operator on real hardware. This could also cost significant time in state-of-the-art tensor compilers~\cite{tvm, zheng2020ansor, ragan2013halide}.

\textbf{Challenges.}
We highlight three main challenges in neural operator synthesis that distinguish it from traditional program synthesis. 
First, with traditional program synthesis, the loop nest (e.g., \cref{list:halide-example}) can be enumerated with bottom-up search, building the coordinate expressions from the atoms such as iterators and constants~\cite{gulwani2017program}.
The main issue with this generic approach is the difficulty of ensuring \emph{high quality} for NN operators, due to the lack of high-level semantics. 
For example, if we fill the indices of \texttt{input} with all 0s, all the other elements would be discarded, which is not at all reasonable.
An NN operator is usually expected to satisfy certain properties, such as differentiability, full utilization of input data elements, etc., so that it can be trained in an NN model and achieve good accuracy. 
Encoding such constraints as inputs to an SMT solver may be possible, but would be too slow when searching over many operator candidates. 

Second, a major issue of exploring the search space is \emph{redundant operators}, which exhibit the same or \emph{similar} semantics and consequently show similar quality and performance. 
For example, in integer arithmetic, there is an identity \texttt{(B*i)\%(B*C)=B*(i\%C)}. Our synthesis needs to skip these equivalent coordinate expressions. 
Moreover, even inequivalent expressions can induce similar computations: considering iterators \texttt{i}, \texttt{j} with domains \texttt{B}, \texttt{K} where $\texttt{B} > \texttt{C} \gg \texttt{K}$, then \texttt{(i+j)/C=i/C} holds for almost every point. 
Traditionally, the redundancy can be handled with term rewrite systems~\cite{newcomb2020verifying} and equality saturation~\cite{eqsat-opt}, but this slows down the search, and cannot prune away inequivalent but similar expressions. 

Third, conventional program synthesis has developed multiple approaches to guide the synthesis with user-provided specifications to eliminate illegal candidates~\cite{gulwani2017program}. 
With neural operators, the only correctness constraint is that the input and output tensor shapes must match with those specified by the model. However, under the aforementioned quality constraints, the domains of coordinate expressions cannot be freely altered to match the input and output shapes, making randomly sampled operators almost always illegal in terms of tensor shapes.
Thus, we need a specially designed novel approach to \emph{guide the synthesis process}.

In summary, to realize practical neural operator synthesis, we must design a framework with the following properties.
\begin{itemize}
    \item \textbf{High quality.} Synthesized operators need to satisfy certain properties (e.g., differentiability, full data utilization), similar to existing NN operators.
    \item \textbf{No redundancy.} Repeated evaluation of operators with the same or similar semantics should be avoided. Particularly, we should not redo the optimizations in existing tensor compilers.
    \item \textbf{Guided search.} The synthesis process should be guided by the input and output tensor shapes to improve the search efficiency.
\end{itemize}

\section{Design Overview}

We propose \sys, an end-to-end, automatic, and efficient framework for neural operator synthesis. 
Given a backbone NN model, \sys is able to synthesize novel linear operators with high quality (for accuracy) and high performance (for speed), which can be drop-in replacements for the original operators (convolution, matmul, etc.) with the same input and output tensor shapes. The model topology and the non-linear activation layers are unaltered.
Specifically, given the input and output tensor shapes, e.g., \texttt{[N,C\textsubscript{in},H,W]} and \texttt{[N,C\textsubscript{out},H,W]} for convolution, or \texttt{[M,K]} and \texttt{[M,N]} for matmul, \sys discovers novel operators that satisfy the accuracy and performance requirements, e.g., best performance with less than 1\% accuracy loss. Note that the tensor shapes are specified as symbolic variables to allow one operator to fulfill different tensor sizes. 
The framework also supports a rich set of user-defined budgets such as FLOPs, memory usage, and number of parameters.

\begin{figure}
    \centering
    \includegraphics[width=0.8\columnwidth]{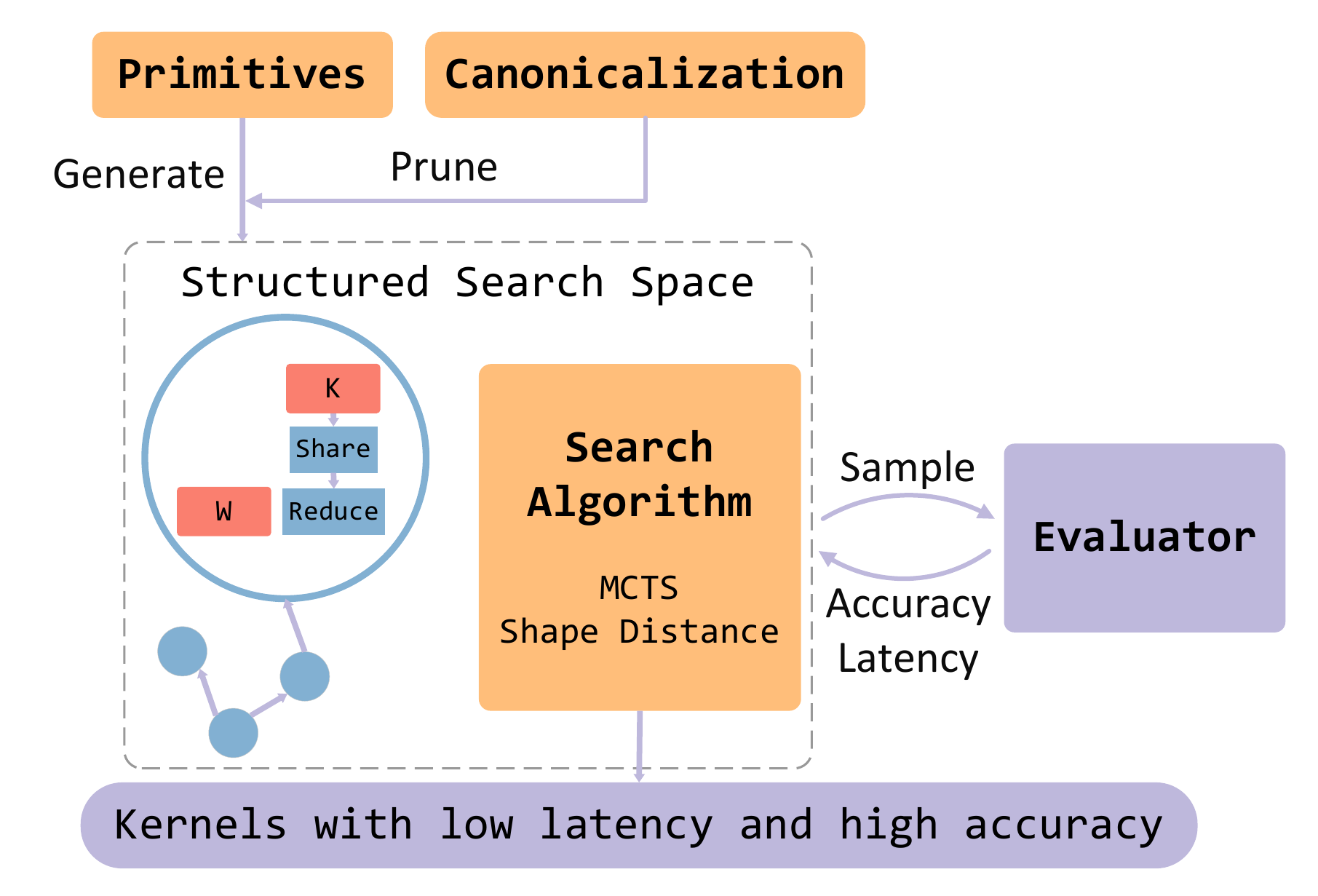}
    \caption{The overall architecture of \sys.}
    \label{fig:overview}
\end{figure}

\begin{algorithm}[t]
\caption{The workflow of \sys.}
\label{alg:overview}
\tablefontsize
\begin{algorithmic}[1]
\Procedure{Search}{$model, d_{max}$}
    \State $operators \gets \Call{ExtractOperators}{model}$
    \State $substs \gets \Call{SynthesizeSubstitutions}{operators, d_{max}}$
    \ForAll{$subst \in \Call{MCTS}{substs}$}
        \State $model^\prime \gets \Call{Replace}{model, operators, subst}$
        \State $accuracy \gets \Call{TrainWithPyTorch}{model^\prime}$
        \If{\Call{IsWithinAccuracyMargin}{$accuracy$}}
            \State $performance \gets \Call{TuneWithTVM}{model^\prime}$
            \State \Output $subst, accuracy, performance$
        \EndIf
    \EndFor
\EndProcedure
\Statex
\Function{SynthesizeSubstitutions}{$operators, d_{max}$}
    \LComment{Search on symbolic shapes, e.g., \texttt{[N, C, H, W]}.}
    \State $input, output \gets \Call{SymbolicShape}{operators}$ \label{alg:overview:symbolic-shape}
    \State $results \gets \{\}$
    \Procedure{Enumerate}{$d, n$}
        \If{\Call{HasMatchingShape}{$n, input$}}
            \State Add $n$ to $results$ if within budgets
        \EndIf
        \If{$d \geq d_{max}$}
            \Return \label{alg:overview:d-max}
        \EndIf
        \ForAll{$n^\prime \in \Call{EnumerateChildren}{n}$}
            \LComment{Backtrack with shape distance.}
            \If{$\Call{ShapeDistance}{n^\prime, input} > d_{max} - d - 1$} \label{alg:overview:shape-dist}
                \State \algorithmiccontinue
            \EndIf
            \State \Call{Enumerate}{$d + 1, n^\prime$}
        \EndFor
    \EndProcedure
    \State \Call{Enumerate}{$0, \Call{RootNode}{output}$}
    \State \Return $results$
\EndFunction
\Statex
\Function{EnumerateChildren}{$n$}
    \State $children \gets \{\}$
    \ForAll{$prim \in \Call{EnumeratePrimitives}{n}$} \label{alg:overview:enum-prim-beg}
        \If{\Call{IsCanonical}{$n, prim$}} \Comment{Canonicalize.} \label{alg:overview:canon}
            \State $children \gets children \cup \{\Call{Add}{n, prim}\}$
        \EndIf
    \EndFor \label{alg:overview:enum-prim-end}
    \State \Return $children$
\EndFunction
\end{algorithmic}

\end{algorithm}

We limit our search in \sys to linear operators. 
First, linear operators are usually the performance bottleneck in NNs, constituting most of the computations, so reducing their complexity can have great gains. 
Second, activation layers like ReLU provide the non-linearity needed in NNs, so we keep them unaltered in the backbone model. Their performance impact is negligible because they can be readily fused into their preceding operators by existing tensor compilers.

\cref{alg:overview} outlines the overall workflow of \sys, which is also illustrated in \cref{fig:overview}. 
\sys relies on a library of \emph{fine-grained primitives} that operate on specific tensor \emph{coordinates}, i.e., dimensions (\cref{sec:primitives}).
A new operator candidate is synthesized by iterative sampling and adding new primitives (\cref{alg:overview} Lines~\ref{alg:overview:enum-prim-beg} to \ref{alg:overview:enum-prim-end}), until reaching a maximum size (Line~\ref{alg:overview:d-max}). 
Compared to directly composing arbitrary coordinate expressions, using these primitives ensures high quality with tensor semantics and enables efficient structured search, while not sacrificing expressiveness. 

Since our primitives are defined on tensor coordinates, we can directly apply \emph{expression simplification and canonicalization} techniques for coordinate expressions to quickly eliminate redundant candidates (\cref{alg:overview} Line~\ref{alg:overview:canon}), enabling efficient search space exploration. Our canonicalization rules not only remove most of the equivalent operators during the search, but can also prune those operators with similar semantics (\cref{sec:pruning}). 

To efficiently discover valid operators, we guide the synthesis flow (\cref{alg:overview} Line~\ref{alg:overview:shape-dist}) using a novel metric of \emph{shape distance}, which is the distance between the current partial operator and a complete operator that has the same input/output shapes as the one in the original model.
We then leverage the intrinsic structure of the search space and formalize the search as a stochastic decision process, in order to apply the Monte Carlo Tree Search algorithm (\cref{sec:search}).
The discovered operators are then fed to the two code generators targeting PyTorch~\cite{ansel2024pytorch} and TVM~\cite{tvm} for accuracy and performance evaluations (\cref{sec:codegen}).

\sys is implemented as a distributed infrastructure, which could leverage multiple GPUs across several server nodes to conduct search, parallelizing the model training required in the accuracy evaluation. \sys has 19K lines of C++ code and 11.5K lines of Python code. \sys is open sourced at \url{https://github.com/tsinghua-ideal/Syno}.

\section{Primitives}
\label{sec:primitives}

\sys adopts a novel approach to synthesize candidate operators from a set of fine-grained primitives, whose semantics are defined with \emph{tensor coordinate expressions} in a bottom-up way, as shown in \cref{tab:primitive-table}. 
Compared with directly enumerating arbitrary raw arithmetic expressions as abstract syntax trees (ASTs) of integer expressions, this allows us to perform synthesis and search with the primitives in a more structured manner to ensure high quality. 

\begin{table*}
\centering\tablefontsize
\caption{
\sys primitives that transform coordinate expressions and their domains in a bottom-up way. For example, \prim{Unfold} combines two coordinates with domains \texttt{N} and \texttt{K}, and obtains an expression of domain \texttt{N} (with out-of-bound elements clipped).
}
\label{tab:primitive-table}
\begin{tabular}{cccclcll}
\toprule
\multicolumn{2}{c}{\textbf{Class}} & \textbf{Primitive} & \textbf{Parameter} & \textbf{Bottom} & & \textbf{Top} & \textbf{Top-Down Semantics} \\
\midrule
\multirow{6}{*}{\makecell{Views}}
 & \multirow{3}{*}{1-to-1}
   & \prim{Split} & - & \texttt{[i, j]: [G, B]} & \textleftarrow & \texttt{[B * i + j]: [G * B]} & Partition into blocks \\
 & & \prim{Merge} & \texttt{B} & \texttt{[i]: [N]} & \textleftarrow & \texttt{[i / B, i \% B]: [N / B, B]} & Flatten two dimensions \\
 & & \prim{Shift} & - & \texttt{[i]: [N]} & \textleftarrow & \texttt{[(i + 1) \% N]: [N]} & Shift along a dimension \\
 \cmidrule{2-8}
 & \multirow{2}{*}{1-to-many}
   & \prim{Expand} & - & \texttt{[i]: [C]} & \textleftarrow & \texttt{[]: []} & Repeat or up-sample \\
 & & \prim{Unfold} & - & \texttt{[i, j]: [N, K]} & \textleftarrow & \texttt{[i + j - K/2]: [N]} & Extract sliding windows \\
 \cmidrule{2-8}
 & \multirow{1}{*}{many-to-1}
   & \prim{Stride} & \texttt{S} & \texttt{[i]: [K]} & \textleftarrow & \texttt{[S * i]: [S * K]} & Strided access \\
\midrule
\multicolumn{2}{c}{\multirow{2}{*}{\makecell{Contractions}}}
   & \prim{Reduce} & \texttt{N} & \texttt{[]: []} & \textleftarrow & \texttt{\textSigma$_\texttt{i}$ [i]: [N]} & Reduce a dimension \\
 & & \prim{Share} & - & \texttt{[i]: [N]} & \textleftarrow & \texttt{([i], [i]): ([N], [N])} & Element-wise product \\
\bottomrule
\end{tabular}
\vspace{10pt}
\end{table*}

\subsection{Structured ASTs}

Synthesizing expressions in a bottom-up way, i.e., first specifying the innermost atoms and then composing them, is common in program synthesis~\cite{gulwani2017program}. This is also a natural choice for NN operators.
As can be seen in \cref{list:halide-example}, each element of the output tensor is calculated through certain arithmetic operations on some elements of the input tensors, which are indexed by some expressions on the indices of the output element. 
Here the output tensor indices, e.g., \texttt{i\_H}, are termed the \emph{output iterators}. They are also the implicitly defined loops in Halide.
The expressions consisting of output iterators and constants are termed \emph{coordinate expressions}. They are used to index the input and output tensors. For example, \texttt{i\_H} is a coordinate expression to index \texttt{out}, and \texttt{i\_H + r\_K\_H - K/2} is also a coordinate expression to index \texttt{input}.
Following the bottom-up approach, we use the output iterators as atom coordinate expressions (the ``bottom''), and enumerate over the diverse combinations of coordinate expressions for the input tensor indices (the ``top'').
By doing so we can synthesize novel operators beyond our current knowledge, and this is the design space we hope to explore.

However, operators synthesized with such straightforward bottom-up enumeration tend to have \emph{low quality}.
For example in \cref{list:halide-example}, if \texttt{i\_Co} were only used in an expression \texttt{i\_Co / 2}, then every two consecutive channels of \texttt{out} would have identical feature maps. This means tensor elements are replicated, and we perform redundant computations. 
To avoid this, we can require that \texttt{i\_Co \% 2} must also be present in the enumerated coordinate expressions. This example inspires us to design a \emph{high-quality} primitive that transforms a coordinate expression \texttt{[i]} with domain \texttt{[N]} to two coordinate expressions \texttt{[i / B, i \% B]} of domains \texttt{[N / B, B]} where \texttt{B} divides \texttt{N}. Formally we write: \texttt{[i]: [N]} ~\textleftarrow~ \texttt{[i / B, i \% B]: [N / B, B]}.
The notation here uses an inverse arrow to point from the ``top'' to the ``bottom'', in order to highlight the dataflow direction from the input tensors to the output.

Furthermore, this \emph{bottom-up primitive} also has \emph{top-down semantics}, namely to flatten a tensor of shape \texttt{[N / B, B]} into \texttt{[N]} by merging the two dimensions. We name it as \prim{Merge}, which is actually a common \emph{tensor view operation}. 
A view is just another way of accessing a tensor. Various arithmetic operations (\texttt{+}, \texttt{*}, \texttt{/}, etc.) on tensor coordinates actually correspond to views. 
For example, the addition of coordinate expressions is equivalent to extracting neighboring elements, which is \prim{Unfold}. 
Similarly, adding a constant is \prim{Shift}, multiplication is \prim{Split} and \prim{Stride}, and discarding an expression is \prim{Expand}. We summarize them in the class of views in \cref{tab:primitive-table}. 
All of them do not discard or replicate elements and thus have high quality, except \prim{Expand} and \prim{Stride}, which could be useful for special cases such as up-sampling and dilated convolution. For semantic completeness, we keep them but limit their occurrences in each synthesized operator.

Aside from coordinate expressions that extract elements from tensors, we need primitives to actually perform computations. For now, we only support linear operations in \sys, so elements from multiple tensors are multiplied and summed up.
The reduction (\texttt{RDom} in \cref{list:halide-example}) can be abstracted to a primitive \prim{Reduce}, which adds a sum reduction loop. 
Meanwhile, a \prim{Share} primitive indexes two tensors with the same coordinate expression, and performs multiplication between the two tensors. 
The top-down semantics of \prim{Reduce} and \prim{Share} are mainly \emph{tensor contraction operations}. A contraction involves combining two tensors along a certain dimension~\cite{greub1978}; e.g., the input channels of input and weight tensors are contracted in \cref{list:halide-example}.

\begin{figure}
    \centering
    \includegraphics[width=0.93\columnwidth]{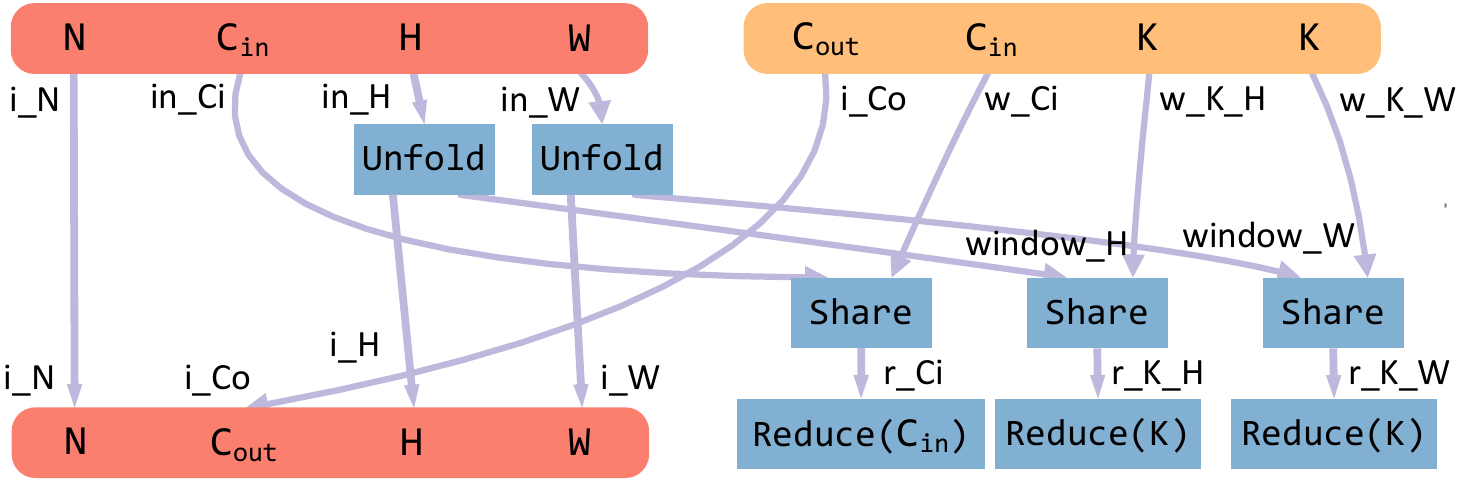}
    \caption{The pGraph of conv2d in \sys. Each edge is a (sub-)coordinate expression. The bottommost orange box is the output tensor of the operator, which comprises the innermost atom coordinate expressions, e.g., \texttt{i\_H: H}. The blue boxes are \sys primitives, each of which transforms the coordinate expressions corresponding to its \emph{out edges} to those corresponding to its \emph{in edges} as specified in \cref{tab:primitive-table}. The topmost orange box is the input tensor to the operator, which comprises the full coordinate expressions. The yellow box is the weight tensor.}
    \label{fig:conv2d_graph}
\end{figure}

With this approach, we propose a structured way of using the \sys primitives to build coordinate expression ASTs for neural operators. An operator composed in this way has a very similar structure to common ASTs, except that instead of trees, expressions are now determined by directed acyclic \emph{primitive graphs} (pGraphs).
\cref{fig:conv2d_graph} shows how to compose a 2D convolution of \cref{list:halide-example} using the \sys primitives. The vertices are the primitives, while the edges are (possibly intermediate) coordinate expressions, and can be evaluated in the same way as we evaluate ASTs.

\subsection{Advantages}
\label{sec:primitives:benefits}

The structured bottom-up primitives in \sys present several advantages. 
First, they ensure \emph{high quality} of synthesized operators, in that they are differentiable~\cite{hu2020jittor} and do not discard input data or replicate data. The only exceptions \prim{Expand} and \prim{Stride} are restrictively used and \prim{Stride} is required to be paired with 1-to-many primitives to ensure the high-quality property. 
Second, they allow \emph{structured search}. With the primitives, similar pGraphs are likely to share a subgraph, which makes the search space highly structured and enables the use of effective search algorithms (\cref{sec:search:algo}). 
Third, the primitives are \emph{expressive}, as they are devised based on the most basic arithmetic operations on coordinate expressions.

\subsection{Semantics and Examples}
\label{sec:primitives:semantics}

To construct an operator from a pGraph like \cref{fig:conv2d_graph}, we evaluate the expressions bottom-up using \cref{tab:primitive-table}, and use them to index the input tensors.
To better illustrate the semantics, we provide examples for several operators: matrix multiplication, pooling, and pixel shuffle in \cref{tab:example-table}, in addition to the convolution example in \cref{fig:conv2d_graph} and \cref{list:halide-example}.

\begin{table*}
\centering\tablefontsize
\caption{
Example operators that can be composed with \sys primitives.
}
\label{tab:example-table}
\begin{tabular}{ccccc}
\toprule
\textbf{PyTorch Operator} & \textbf{Constituent Primitives} & \textbf{pGraph} & \textbf{Halide Code} \\
\midrule
\texttt{mm(input, weight)} &
\makebox[0.34\textwidth][l]{
$\begin{aligned}
& \mathtt{[i, j]: [M, N]} \\
\alignedleftarrow{\prim{Reduce}\mathtt{(K)}} & \mathtt{[i, j, r\_K]: [M, N, K]} \\
\mathmakebox[\arrow]{\stackrel[\mathtt{(Match)}]{\prim{Share}}{\longleftarrow}} & \mathtt{([i, r\_K], [r\_K, j]): ([M, K], [K, N])}
\end{aligned}$} &
\raisebox{-0.5\height}{\includegraphics[width=0.125\textwidth]{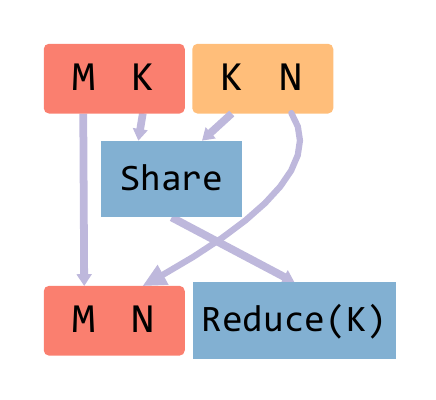}} &
\begin{lstlisting}
auto [r_K] = RDom(0, K);
out(i, j) +=
  input(i, r_K)
  * weight(r_K, j);
\end{lstlisting} \\
\midrule
\texttt{nn.AvgPool1d(s)(input)} &
\makebox[0.34\textwidth][l]{
$\begin{aligned}
& \mathtt{[i]: [s^{-1}*H]} \\
\alignedleftarrow{\prim{Reduce}\mathtt{(s)}} & \mathtt{[i, r\_s]: [s^{-1}*H, s]} \\
\alignedleftarrow{\prim{Split}} & \mathtt{[s*i+r\_s]: [H]}
\end{aligned}$} &
\raisebox{-0.5\height}{\includegraphics[width=0.1\textwidth]{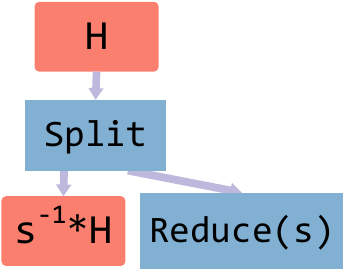}} &
\begin{lstlisting}
auto [r_s] = RDom(0, s);
out(i) +=
  input(s*i+r_s);
\end{lstlisting} \\
\midrule
\texttt{nn.PixelShuffle(B)(input)} &
\makebox[0.34\textwidth][l]{
$\begin{aligned}
& \mathtt{[i]: [H]} \\
\alignedleftarrow{\prim{Merge}\mathtt{(B)}} & \mathtt{[i/B, i\%B]: [B^{-1}*H, B]} \\
\alignedleftarrow{\prim{Split}} & \mathtt{[(H/B)*(i\%B)+i/B]: [H]}
\end{aligned}$} &
\raisebox{-0.5\height}{\includegraphics[width=0.04\textwidth]{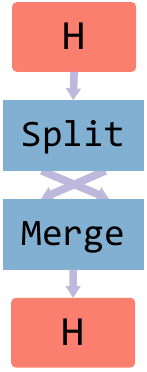}} &
\begin{lstlisting}
out(i) =
  input((H/B)*(i%B)+i/B);
\end{lstlisting} \\
\bottomrule
\end{tabular}
\end{table*}

For example, to obtain a PyTorch operator \texttt{torch.mm} for matrix multiplication, \sys starts with the \emph{bottom} coordinate expressions that index the output tensor, i.e., \texttt{[i. j]: [M, N]} for \texttt{mm}. 
It gradually applies the primitives \prim{Reduce}\texttt{(K)} and \prim{Share}, to compose a valid pGraph.
The \emph{top} coordinate expressions can be used to directly index the input tensors \texttt{input(i, r\_K)} and \texttt{weight(r\_K, j)}.
More specifically, after \prim{Reduce}\texttt{(K)} is applied to introduce a reduction, \prim{Share} is applied to assign one \texttt{r\_K: K} to index the input, and one \texttt{r\_K: K} to index the weight.
A subtle detail here is that, without further restriction, \texttt{i: M} and \texttt{j: N} can be used to index either the input or weight tensor, but here we want exactly \texttt{j: N} to index the weight.
So an implicit \texttt{Match} step is done along with \prim{Share} to assign \texttt{j: N} to the weight tensor.
It tracks all the coordinate expressions to be assigned to the new weight tensor created by a \prim{Share}, and is always applied right after the \prim{Share}. Thus we treat \texttt{Match} as an implementation detail of \prim{Share}, and do not place too much emphasis on it.
The other two example operators and the convolution in \cref{fig:conv2d_graph} are similar.

\subsection{Design Details}
\label{sec:primitives:details}

To match operators with different concrete input/output tensor shapes, and to support additional parameter variables in some primitives (e.g., \prim{Merge} needs a factor \texttt{B}), \sys uses \emph{symbolic shapes} when synthesizing operators. 
We further split the symbols into two classes. 
\emph{Primary variables} are for input/output dimensions, e.g., \texttt{C\textsubscript{out}}, \texttt{H}. They are relatively large and thus are not allowed to appear in the denominator of a coordinate expression.
\emph{Coefficient variables} are only introduced by primitives, and are relatively small and allowed to appear in denominators.
When enumerating the applicable primitives on a partial pGraph, the primitive parameters are represented by monomials of primary variables and coefficient variables, with the degrees (i.e., powers) limited within a user-specified range.
\sys replaces the variables with concrete sizes at code generation. 

In the current prototype of \sys, we only consider operators that process a single input tensor (not including weights) and produce a single output tensor, and disallow multiple uses of the same (input or intermediate) tensors such as residual links~\cite{resnet}. 
This restriction seems strict, but in fact existing operator types like convolution, matmul, and pooling all satisfy it. 
We argue that the lost flexibility is usually more critical at the full model graph level rather than at the operator level. Our operators can still be plugged into arbitrary model topologies including ResNet~\cite{resnet}, where the residual links are realized outside the operators. 
We plan to extend \sys to support multiple input tensors in the future.

\begin{figure*}
    \centering
    \includegraphics[width=0.85\textwidth]{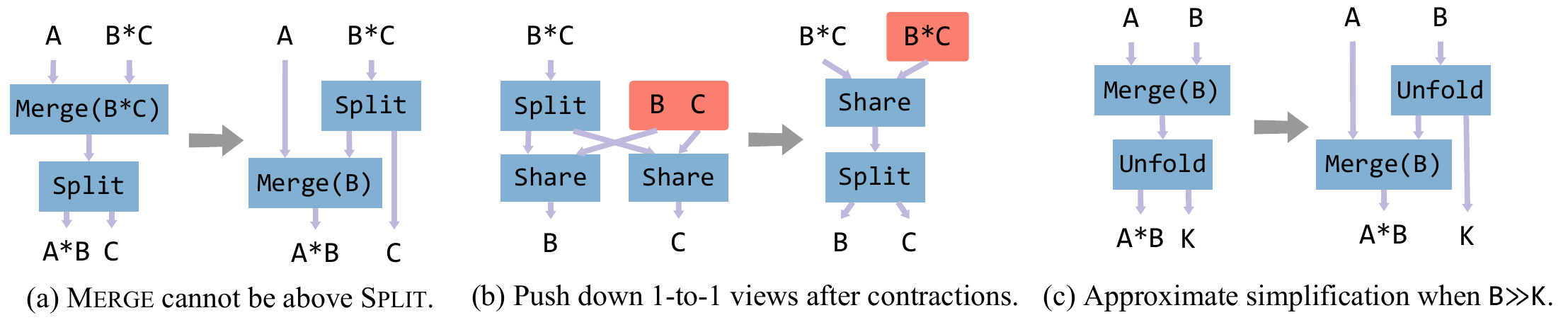}
    \caption{Examples of some canonicalization rules used in \sys.}
    \label{fig:canon}
\end{figure*}

\section{Canonicalization}
\label{sec:pruning}

The design space of synthesizing operators from our primitives is extremely large, with a lot of redundant operator constructs, especially those that can be readily discovered by tensor compilers. 
Take the partial pGraph in \cref{fig:canon}(a) as an example. On the left side, the topmost coordinate expressions are given by \texttt{[i, j]: [A*B, C]} $\stackrel{\prim{Split}}{\longleftarrow}$ \texttt{[C*i+j]: [A*B*C]} $\stackrel{\prim{Merge}\mathtt{(\texttt{B*C})}}{\longleftarrow}$ \texttt{[(C*i+j)/(B*C), (C*i+j)\%(B*C)]: [A, B*C]}.
However, this simplifies to \texttt{[i/B, C*(i\%B)+j]}, corresponding to the right side \texttt{[i, j]: [A*B, C]} $\stackrel{\prim{Merge}\mathtt{(\texttt{B})}}{\longleftarrow}$ \texttt{[i/B, i\%B, j]: [A, B, C]} $\stackrel{\prim{Split}}{\longleftarrow}$ \texttt{[i/B, C*(i\%B)+j]: [A, B*C]}.

To improve search efficiency, \sys uses a set of \emph{canonicalization rules} to filter out uncanonical redundant candidates on the fly when new primitives are added to partial pGraphs (\textsc{IsCanonical} in \cref{alg:overview} Line~\ref{alg:overview:canon}).
\sys does \emph{not} aim to eliminate \emph{all} redundancies, which is highly challenging, if not impossible, considering the rich primitive semantics. Also, comprehensive canonicalization checks are extremely expensive and sometimes undecidable~\cite{newcomb2020verifying}. 
On the other hand, semantically similar models have similar quality and performance, so \sys supports canonicalization rules to mark only one operator as canonical among the many ones in the class that have \emph{similar computation results}.
We also note that the canonicalization rules in \sys are easily \emph{extensible}. Developers can define new rules and plug them into the framework.

\textbf{Contractions.}
Since weight tensors can be arbitrarily reshaped offline, there is no need to apply views to weights.
Thus weight coordinates are directly used in \prim{Share}s for contractions\footnote{This also implies that the coordinate expressions assigned to weights by the \texttt{Match} step mentioned in \cref{sec:primitives:semantics} can be removed from the set of coordinate expressions that can be further transformed and matched against the desired topmost shape, simplifying the synthesis algorithm in \cref{sec:search:synth}.}.
Moreover, we always put weight coordinates as the right-hand-side inputs of the symmetric \prim{Share}s.

\textbf{Between views and contractions.}
We enforce a canonical order between views and contractions.
The 1-to-1 views do not involve actual computations so they can be freely swapped with contractions. We thus \emph{push down} all 1-to-1 views after contractions, as in \cref{fig:canon}(b). 

For the others, we apply rules to avoid doing futile work. 
For example, we disallow combining \prim{Expand} and \prim{Reduce} because this only changes a multiplier of the result;
\prim{Unfold} allows at most one output coordinate to be \prim{Reduce}d.

\textbf{Between two views.}
Most redundancies exist between views because of the many primitive types. 
The key to their canonicalization is \emph{to apply expression simplification techniques}. In many tensor compilers such as Halide~\cite{ragan2013halide} and TVM~\cite{tvm}, expressions are simplified before analysis and lowering.
In \sys, as our primitive definitions are based on coordinate expressions, we can similarly simplify the coordinate expressions corresponding to the (sub)pGraph consisting of view primitives, and the fully simplified expression gives the canonical form.
For example, the aforementioned \cref{fig:canon}(a) is one such example, where the right side is simplified and thus canonical, corresponding to the rule that a \prim{Merge} cannot be above a \prim{Split}.

We design expression simplification in \sys by referring to Halide's term rewrite system (TRS)~\cite{newcomb2020verifying}. TRS sequentially substitutes the terms in an AST from bottom to top with pattern matching in expressions, in order to obtain the simplified form~\cite{newcomb2020verifying}. 
In \sys, rather than actually rewriting the pGraph, canonicalization is applied on-the-fly when we add new primitives to the partial pGraph, by discarding candidates that create uncanonical forms (so we need not worry about the termination of rewrites).
In our pGraph, each coordinate (edge) is like an AST node. We treat the bottom outputs of a subgraph as wildcards and match the top input expressions against the patterns.
Again look at \cref{fig:canon}(a). The bottom outputs are marked as \texttt{[\#0, \#1]}. 
Then the substitution can be formulated as \texttt{[(C*\#0+\#1)/(B*C), (C*\#0+\#1)\%(B*C)]} \texttt{->} \texttt{[\#0/B, C*(\#0\%B)+\#1]}, which is a pattern-matching-based rewrite rule. 
To choose the canonical (i.e., ``simplest'') form among equivalent expressions, we empirically define simplicity as removing parentheses as much as possible by applying distribution laws of multiplication, division, and modulo. 
We can see \cref{fig:canon}(a) removes one level of parentheses. 
Following this approach, we derive a series of rewriting rules involving multiple primitives.

In addition, it is better to not just canonicalize semantically equivalent subgraphs, but also eliminate candidates with only slightly different semantics, so that a wider range of semantics can be explored with fewer samples.
In \cref{fig:canon}(c), on the left side, with \texttt{[i, j]: [A*B, K]} as the output coordinates, the inputs are \texttt{[(i+j-K/2)/B, (i+j-K/2)\%B]: [A, B]}.
If \texttt{B} is much larger than \texttt{K} as in most convolutions\footnote{While we are using symbolic shapes during synthesis (\cref{sec:primitives:details}), we also extract all possible concrete values for each symbolic shape from the input backbone NN model. Symbolic $\texttt{B} \gg \texttt{K}$ is true if for every valuation of \texttt{B} and \texttt{K} we have $\texttt{B} \gg \texttt{K}$.}, then \texttt{j-K/2} is much less than \texttt{B}. So we can simplify the expressions to \texttt{[i/B+j-K/2, i\%B+j-K/2]} as the right side, which is equal to the left side at almost every point.
Other similar rules are devised based on the principle of removing parentheses, and these approximately equivalent rules can also be implemented as TRS-based rules as above.
They effectively enable us to only synthesize operators that are significantly distinct.

\section{Guided Search}
\label{sec:search}

We next describe the overall synthesis and search process in \sys. 
\cref{sec:search:synth} discusses the bottom-up synthesis approach. A critical challenge is how to ensure the exact match of the input/output tensor dimensions with the given specification. We propose a novel concept of shape distance to guide the search.
\cref{sec:search:algo} explains our specific search algorithm based on Monte Carlo Tree Search (MCTS)~\cite{coulom2007mcts, gaudel2010fuse, hsu2020mcts}.

\subsection{Bottom-Up Synthesis with Shape Distance}
\label{sec:search:synth}

As mentioned in \cref{sec:primitives}, \sys performs bottom-up synthesis, starting from the output coordinates and iteratively applying sampled primitives for a limited number of steps. 
For example, as a subgraph of \cref{fig:conv2d_graph}, from the output \texttt{[i\_H]: [H]}, we can get \texttt{[i\_H]: [H]} $\stackrel{\prim{Reduce}\mathtt{(\texttt{K})}}{\longleftarrow}$ \texttt{[i\_H, r\_K\_H]: [H, K]} $\stackrel{\prim{Share}}{\longleftarrow}$ \texttt{([i\_H, r\_K\_H], [r\_K\_H]): ([H, K], [K])} $\stackrel{\prim{Unfold}}{\longleftarrow}$ \texttt{([i\_H + r\_K\_H - K / 2], [r\_K\_H]): ([H], [K])}. The weight (\texttt{[K]} here) does not need to be transformed further (\cref{sec:pruning}), so we use the term \emph{shape} to refer to the shape of the first tensor (data input tensor), which is \texttt{[H]} in this case.

The data tensor shape of a complete pGraph should match exactly with the \emph{desired shape} (the shape of $input$ in \cref{alg:overview} Line~\ref{alg:overview:symbolic-shape}).
While synthesizing with primitives ensures high quality, it also becomes hard to control the \emph{dimensions} (sizes of tensor coordinates) after applying primitives on a partial pGraph.
Ideally, after flexibly exploring various primitives, when the partial pGraph gets close to its maximum size limit, the last few primitives need to move towards exactly matching with the desired dimensions.
For example, if the shape of the current partial pGraph is \texttt{[C\textsubscript{in}, s\textsuperscript{-1}*H, s*W, k]}, we can apply \texttt{[C\textsubscript{in}, s\textsuperscript{-1}*H, s*W, k]} $\stackrel{\prim{Merge}}{\longleftarrow}$ \texttt{[C\textsubscript{in}, s\textsuperscript{-1}*H, s, W, k]} $\stackrel{\prim{Split}\mathtt{(s)}}{\longleftarrow}$ \texttt{[C\textsubscript{in}, H, W, k]} $\stackrel{\prim{Unfold}}{\longleftarrow}$ \texttt{[C\textsubscript{in}, H, W]}.

We propose a novel metric named \emph{shape distance} as the minimum number of required primitives added onto the current pGraph to reach the desired shape. 
In the above example, the shape distance of \texttt{[C\textsubscript{in}, s\textsuperscript{-1}*H, s*W, k]} is $3$. 
If the remaining allowed number of primitives is less than the shape distance, we can immediately terminate the current pGraph and backtrack (\cref{alg:overview} Line~\ref{alg:overview:shape-dist}). This avoids deviating too far from the desired dimensions.

We design a systematic method in \sys to compute the shape distance between the current shape and the \emph{desired shape}.
We first divide the dimensions in the two shapes into \emph{reshape groups}, where future primitives are only applied to the dimensions within each group to match them, but not across groups. 
In the above example, we can have three reshape groups, as \texttt{\{C\textsubscript{in}\} <- \{C\textsubscript{in}\}}, \texttt{\{s\textsuperscript{-1}*H, s*W\} <- \{H, W\}}, \texttt{\{k\} <- \{\}}.
Reshape groups can be decided by comparing the primary variables in the coordinate expressions. 
When there exist multiple possible grouping schemes (but usually only a few), we enumerate all and find the least distance.

We then compute the distance within each reshape group.
We identify the \emph{helpful primitives} that will help in shape matching: reshape primitives (i.e., \prim{Merge}, \prim{Split}) which regroup dimensions, and 1-to-many primitives (i.e. \prim{Unfold}, \prim{Expand}) which eliminate dimensions.
If the left-hand side and the right-hand side of the reshape group have the same size of domains, e.g., \texttt{\{s\textsuperscript{-1}*H, s*W\}} and \texttt{\{H, W\}} have domains of \texttt{H*W}, then we only need to regroup dimensions, using \prim{Split} and \prim{Merge}. In this case we only need 2 steps: \texttt{[s\textsuperscript{-1}*H, s*W]} $\stackrel{\prim{Merge}}{\longleftarrow}$ \texttt{[s\textsuperscript{-1}*H, s, W]} $\stackrel{\prim{Split}\mathtt{(s)}}{\longleftarrow}$ \texttt{[H, W]}. We can prove a generalized conclusion of $\#lhs+\#rhs-2$ steps, where $\#lhs$ and $\#rhs$ are the numbers of dimensions in the left-hand and right-hand sides of the reshapes (both are 2 in this example).
On the other hand, if the two sides have different sizes of domains, then at least one 1-to-many primitive is required, counting as $1$ extra step.
We sum up the bounds ($\#lhs+\#rhs-2$) of all the reshape groups, adding $1$ if the domain of the current shape is different from the desired shape, and use it as an upper bound for shape distance.
Then, all grouping schemes are enumerated to find the minimum of the upper bounds, which yields the final shape distance. 

When the desired shape involves repeated dimensions, e.g., \texttt{[C\textsubscript{in}, H, H]} for square images, we enumerate all possible permutations, allowing tensor transpose during the final matching.

\subsection{MCTS-Based Search}
\label{sec:search:algo}

Our search algorithm is based on MCTS~\cite{coulom2007mcts, gaudel2010fuse, hsu2020mcts}.
We formulate our search problem as a Markov decision process, where we transit from one partial pGraph to another in the search space, with the action space being the primitives. The final states are complete pGraphs. 
The optimization goal is operators with both high accuracy and high inference speed.
As the FLOPs of operators are much easier to compute than the inference accuracy which requires extensive training, we set a hard upper limit for FLOPs and use accuracy as the reward for MCTS to guide it to learn how to find expressive operators within a given FLOPs budget. 
We record all MCTS samples and filter out operators with bad accuracies to obtain the final result. 

\section{Code Generation}
\label{sec:codegen}

We implement two code generators for accuracy and speed evaluations.
First, a \emph{PyTorch code generator} is built to make use of the already highly-tuned operator libraries for training. Using the top-down semantics, each view primitive is lowered to its counterpart in PyTorch, and each contraction primitive is lowered to an einsum~\cite{rogozhnikov2021einops} expression, which is a general method for performing tensor contractions. The primitives are lowered in topological order to ensure that dependencies are satisfied.
We further use TorchInductor~\cite{ansel2024pytorch} for compile-time optimizations such as fusion and tiling.

However, PyTorch and TorchInductor are mainly optimized for existing workloads and tuned on a limited set of operators such as convolution and matrix multiplication. To better support the novel opeartors discovered by \sys, we further build a \emph{TVM TE (Tensor Expression) code generator}, to utilize the more general-purpose compiler, TVM~\cite{tvm}.
It follows the bottom-up semantics to evaluate all coordinate expressions according to the pGraph, and leverages TVM for extensive compiler optimizations on specific hardware, e.g., our mobile CPUs and GPUs in \cref{sec:eval}. The TVM TE syntax is very close to that of Halide as we mentioned earlier, so we skip the technical details.

\begin{figure}
    \centering
    \includegraphics[width=0.7\columnwidth]{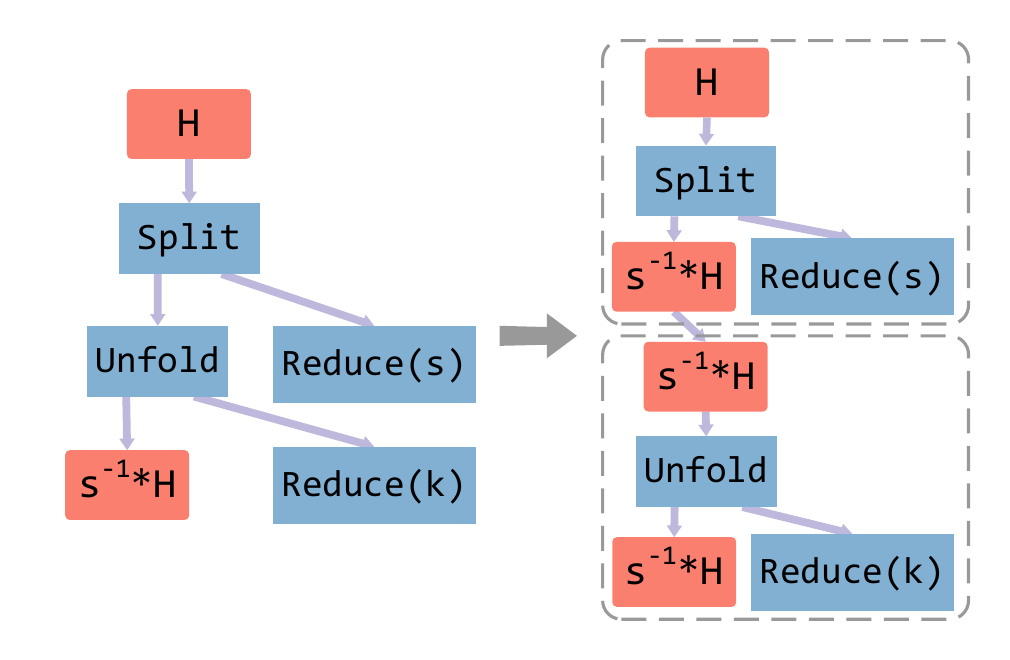}
    \caption{An example of the materialized reduction optimization in \sys.}
    \label{fig:codegen-rfactor}
\end{figure}

Some optimization passes unique to \sys are designed. An important one aims to automatically insert intermediate stages (materializations) to eliminate redundant computations. 
Consider the example in \cref{fig:codegen-rfactor}. A trivial code generator creates a loop nest of \texttt{(H/s)*k*s} iterations computing
\texttt{Y[i] = $\sum_{\texttt{i$_\texttt{k}$}}\sum_{\texttt{i$_\texttt{s}$}}$X[i + i$_\texttt{k}$ - k/2 + s*i$_\texttt{s}$]}
as on the left side. But this is mathematically equivalent to
\texttt{Z[i$^\prime$] = $\sum_{\texttt{i$_\texttt{s}$}}$X[i$^\prime$ + s*i$_\texttt{s}$], Y[i] = $\sum_{\texttt{i$_\texttt{k}$}}$Z[i + i$_\texttt{k}$ - k/2]},
which corresponds to the partitioned subgraphs on the right. 
By doing so we reduce the FLOPs from \texttt{k*H} to \texttt{(1+k/s)*H}. 

Generally speaking, the FLOPs depend only on the output iterators and the \prim{Reduce}s, which are the spatial loops and reduction loops in the loop nest. The number of iterations is their product. In the case of 1-to-many primitives like \prim{Unfold}, the output dimensions are increased, so if we perform any \prim{Reduce} after this, FLOPs are unnecessarily increased because we are evaluating \texttt{k} copies for each element. 

This issue is unique to the \sys IR. To deal with it, we propose an optimization named \emph{materialized reduction}, which materializes the bottom (output tensor) of a sub-pGraph that performs reductions.
We enumerate the order of performing reductions, i.e., the order of lowering each \prim{Reduce}. If a \prim{Reduce} is lowered, only the primitives that can reach that \prim{Reduce} are required to be lowered. In the example, the \prim{Split} and the \prim{Unfold} primitives can both reach the bottom \prim{Reduce}, but only the 1-to-many \prim{Unfold} cannot reach the upper \prim{Reduce}. So the upper \prim{Reduce} is prioritized to form a sub-pGraph, materializing the bottom.

\section{Evaluation}
\label{sec:eval}

\subsection{Experimental Setups}

\textbf{Hardware configurations.}
Our operator search and accuracy validation are done on a cluster with NVIDIA A100 GPUs. 
For performance in edge-device inference scenarios, we test the end-to-end latency on NVIDIA Jetson Orin Nano 8 GB, which features a 6-core Arm Cortex-A78AE mobile CPU and a 1024-core NVIDIA Ampere GPU with 32 tensor cores. 
For performance on server-grade GPUs, we evaluate the end-to-end latency on an NVIDIA A100 GPU.
In summary, we evaluate performance on three platforms: (1) mobile CPU, (2) mobile GPU, and (3) A100.

\textbf{Compilers.}
To demonstrate the orthogonality of \sys to tensor compilers and its wide applicability, we evaluate on two compilers: (1) TVM MetaSchedule~\cite{shao2022tensor}, a state-of-the-art tuning-based tensor compiler widely adopted by the research community; (2) TorchInductor, the default \texttt{torch.compile} backend of PyTorch 2~\cite{ansel2024pytorch}, widely adopted by the industry, with its \texttt{max-autotune} mode enabled.

\textbf{Workloads.}
We mainly focus on vision tasks with five popular vision NNs: ResNet-18~\cite{resnet}, ResNet-34~\cite{resnet}, DenseNet-121~\cite{densenet}, ResNeXt-29-2x64D~\cite{resnext}, and EfficientNet-V2-S~\cite{efficientnetv2}. We aim to substitute all standard convolutions in them.
To prove the wide adaptability of \sys, we also test GPT-2~\cite{gpt2} (117M parameters with 12 layers, 12 heads, and 768 embedding dimensions) by substituting its QKV projections.

\textbf{Baselines.}
We use three baselines for the comparison of vision tasks.
The main baseline is the original models with standard convolutions. We target the latency-accuracy tradeoff, so we expect to reduce the end-to-end latency at the cost of minor accuracy degradation.
\citet{turner2021neural} (labeled as NAS-PTE) are the first to introduce loop-level transformations into NAS, and $\alpha$NAS~\cite{jin2022neural} is the first attempt to apply program synthesis for NAS albeit at a coarse granularity.
Because the search and tuning methods of NAS-PTE are not open-source, we compare with their operators on individual layers instead of full models. 
$\alpha$NAS is neither open-source nor provides inference performance data, so we only compare against their FLOPs and training speedups reported in the original paper. 
For GPT-2, we evaluate the training speed relative to the original model. 

\textbf{Datasets and training configurations.}
ImageNet~\cite{imagenet} is unsuitable for direct search because of its large size, so we use the smaller yet still challenging CIFAR-100~\cite{cifar100} as the proxy dataset.
Specifically, during the search \sys trains the NN model using each candidate operator for 100 epochs on CIFAR-100. The selected best operators are then fully trained on ImageNet for 90 epochs for accuracy and performance evaluations. We scale the CIFAR-100 images to the same size as ImageNet to ensure the same inference performance.
For GPT-2, we employ the language perplexity (PPL) metric on the lm1b benchmark~\cite{lm1b}. 
The data type for both training and inference is FP32. 
The training hyperparameters for the optimizer and learning rate scheduler are dataset-dependent to ensure reasonable accuracy, but they are not heavily tuned. 

\textbf{Computation cost.}
Training a model on CIFAR-100 for 100 epochs takes two to three hours. 
In our experiments, we terminate early when the accuracy is not as high as expected, thereby reducing the average evaluation computation cost to 0.1 GPU hours per sample. 
We spend roughly 300 GPU hours per model.

\subsection{Results on Vision Tasks}

For vision tasks, we search for the fastest operators in each model with less than 1\% accuracy loss, a commonly used threshold. We separately target both CPUs and GPUs.

\begin{figure}
    \centering
    \includegraphics[width=\columnwidth]{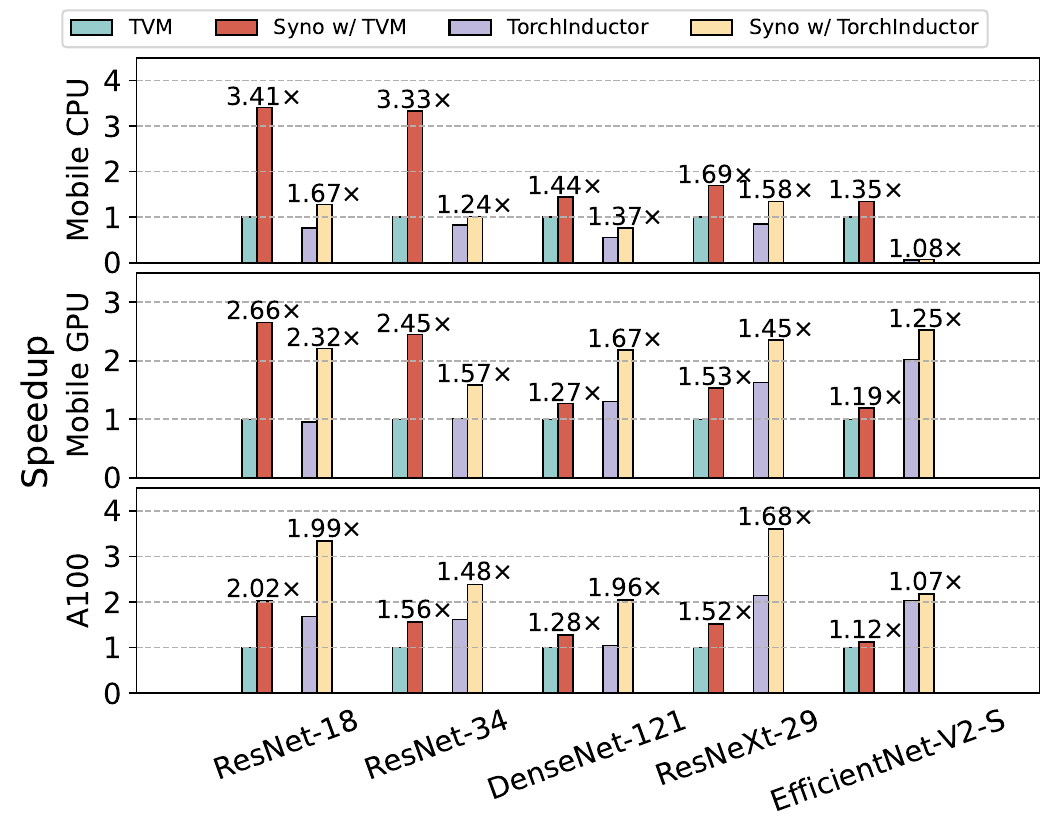}
    \caption{End-to-end performance speedup of \sys on CIFAR-100. The bars of each model are normalized to TVM for direct comparison across different compilers.}
    \label{fig:cifar100-perf}
\end{figure}

\begin{figure*}
    \centering
    \includegraphics[width=\textwidth]{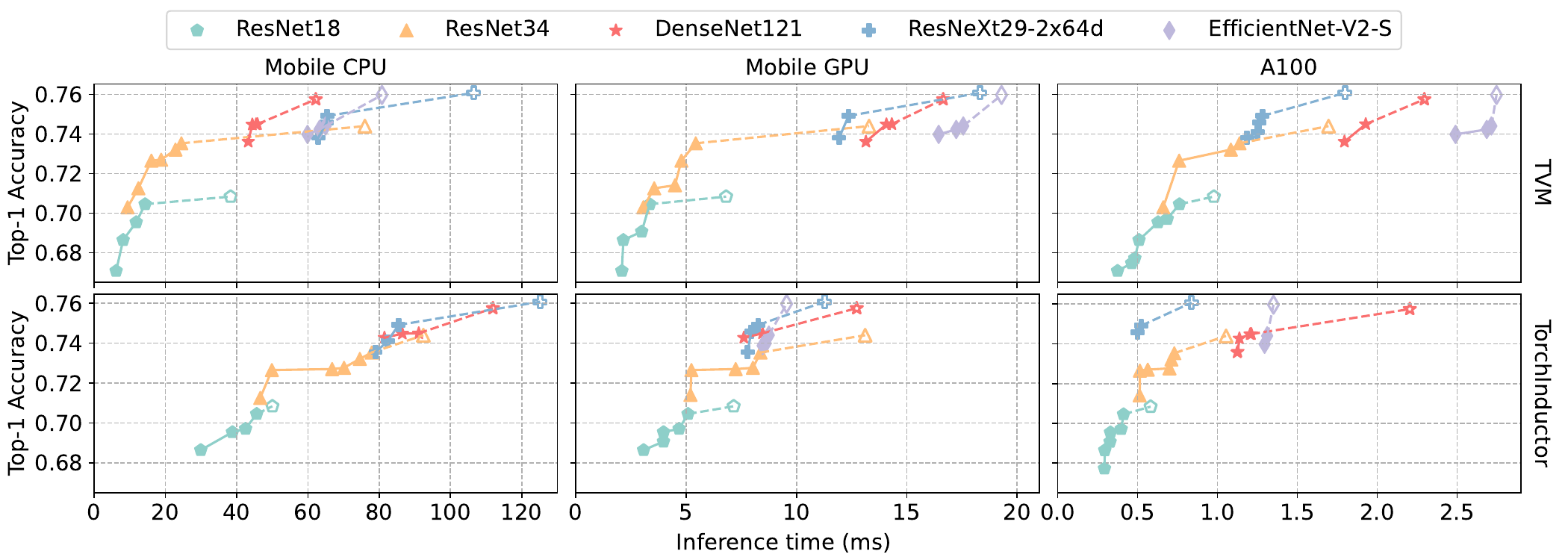}
    \caption{Pareto optimal curves of accuracy vs. inference time between the original and the \sys-optimized models on ImageNet. For each model, the hollow point is the baseline, and the connected solid points are discovered by \sys.}
    \label{fig:imagenet-result}
\end{figure*}

\textbf{CIFAR-100 results.}
\cref{fig:cifar100-perf} shows the best operators we find in terms of inference latency within the accuracy loss limit. 
On the mobile CPU, the mobile GPU, and A100, respectively, \sys achieves $2.06\times$, $1.72\times$, and $1.47\times$ end-to-end inference speedups over the original models on average (geomean) when compiled with TVM, and $1.37\times$, $1.62\times$, and $1.60\times$ when compiled with TorchInductor.
\sys performs better on traditional NNs like ResNet-18 with the discovered novel operators. 
Even for NAS-optimized models such as EfficientNet-V2, \sys can still achieve a performance gain up to $1.35\times$.
We perform more detailed analysis on the benefits of our newly discovered operators later.

It is interesting to compare the two compiler backends, TVM and TorchInductor. 
We find that for the FP32 data type we use, TVM cannot make use of the tensor cores (which requires TF32), so it is generally slower than TorchInductor on GPUs. 
However, TVM tunes over a much larger search space and performs code generation for every operator, whereas TorchInductor can only select from several templates and would conservatively fall back to PyTorch ATen kernels for small GPUs or if the few templates are too slow. Therefore TorchInductor yields more unstable performance than TVM. 
For instance in \cref{fig:cifar100-perf}, TorchInductor performs poorly on EfficientNet-V2-S when using the mobile CPU. Profiling indicates that TorchInductor falls back to use ATen grouped convolution in most cases, which has terrible performance for the many depth-wise convolutions in this model.

\textbf{ImageNet results.}
For every model, we select some discovered operators that have comparable accuracy with the baseline and re-evaluate them on ImageNet. We plot the Pareto optimal curves of accuracy vs. inference time in \cref{fig:imagenet-result}. 
Most of our operators exhibit a minor 1\% to 2\% accuracy loss, while enabling up to $4.73\times$ and $1.94\times$ speedups when compiled with TVM and TorchInductor, respectively. If more accuracy loss is acceptable, then going along the Pareto curves further boosts performance.

We highlight a comparison between our optimized ResNet-34 and the baseline ResNet-18. 
Replacing the standard convolutions with our operators in ResNet-34 results in a model with \emph{both} higher accuracy and better inference time than the ResNet-18 baseline. 
This observation implies a potentially promising direction to extend \sys to accuracy-preserving NN optimization: users can stack more layers and then compress the model with \sys, which might result in better accuracy and lower latency at the same time. 

\begin{figure}
    \centering
    \includegraphics[width=\columnwidth]{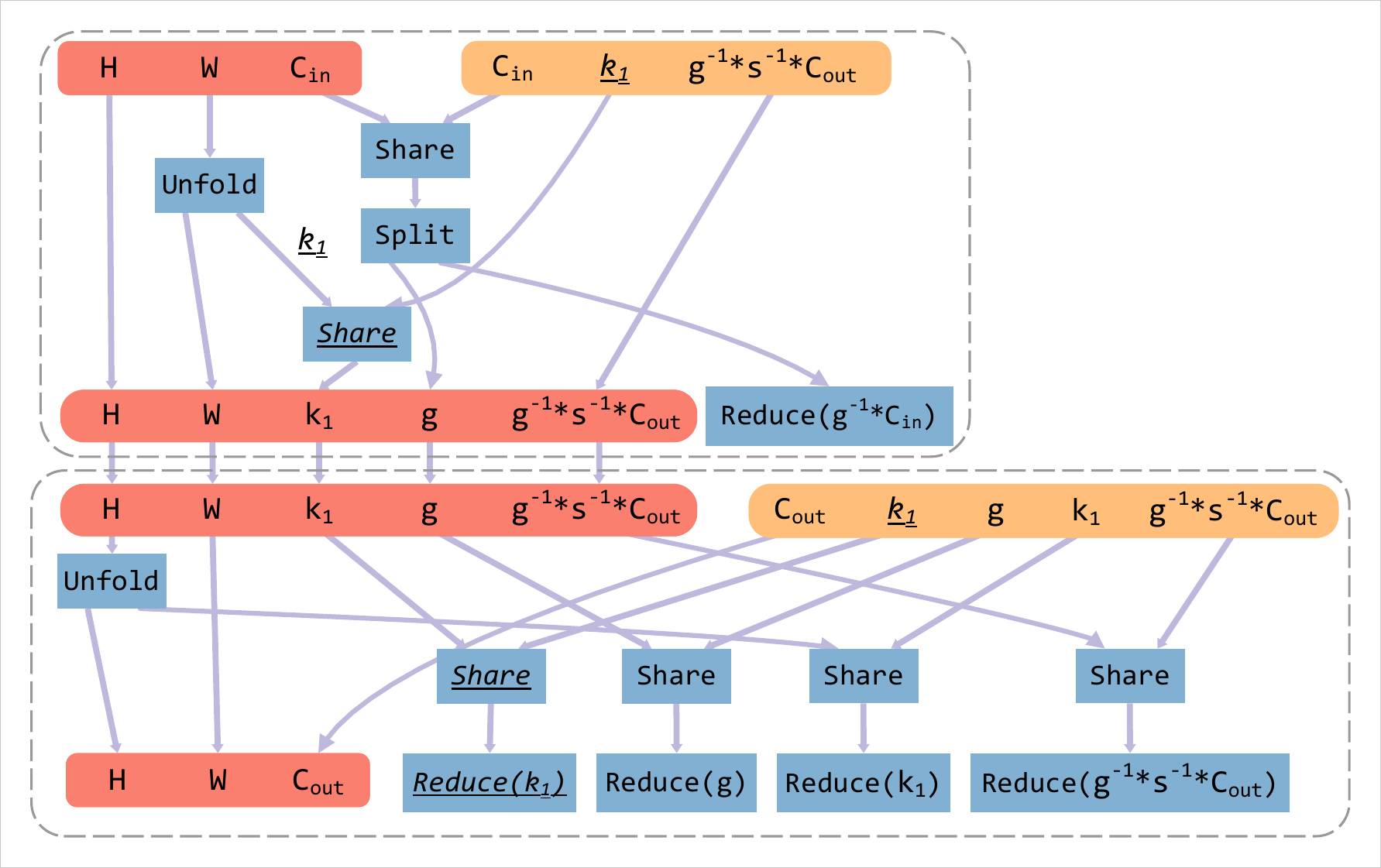}
    \caption{\textsc{Operator 1} discovered by \sys.}
    \label{fig:kernel1-dfg}
\end{figure}

\begin{listing}[ht]
\begin{minted}[fontsize=\footnotesize]{python}
def __init__(self):
  self.w1 = randn([C_out//g//s, C_in, k_1])
  self.w2 = randn([C_out, k_1*k_1*C_out//s])
def forward(self, x):
  N, C_in, H, W = x.shape
  x = nn.functional.unfold(x, [1, k_1], padding=[0, k_1//2])
  # x: [N, C_in*k_1, H, W]
  x = reshape(x, [N, C_in, k_1, H, W])
  x = einsum("nckhw, dck -> ndckhw", x, self.w1)
  # x: [N, C_out//g//s, C_in, k_1, H, W]
  x = reshape(x, [N, C_out//g//s, g, C_in//g, k_1, H, W])
  x = sum(x, 3) # x: [N, C_out//g//s, g, k_1, H, W]
  x = reshape(x, [N, k_1*C_out//s, H, W])
  x = nn.functional.unfold(x, [k_1, 1], padding=[k_1//2, 0])
  # x: [N, k_1*k_1*C_out//s, H, W]
  x = einsum("nchw, dc -> ndhw", x, self.w2)
  return x # x: [N, C_out, H, W]
\end{minted}
\caption{PyTorch code for \textsc{Operator 1}.}
\label{list:operator_1_torch}
\end{listing}

\textbf{Case studies.}
Among all the operators discovered, we find two convolution-like operators with outstanding accuracy and inference performance.

\textsc{Operator 1} shown in \cref{fig:kernel1-dfg} achieves $2.68\times$, $2.04\times$, and $1.28\times$ speedups on the three hardware platforms, with less than 1\% ImageNet accuracy degradation. Its PyTorch code is shown in \cref{list:operator_1_torch}.
After the materialized reduction optimization during code generation (\cref{sec:codegen}), it becomes a stack of two stages \emph{similar to} 1D and 2D grouped convolutions, but is \emph{not} expressible in NAS.
NAS can only sample traditional (grouped) convolutions, which always perform contractions between \prim{Unfold}ed windows of spatial dimensions and weights (the standard \textSigma\textsubscript{\texttt{j}} \texttt{X[i + j - K / 2] * W[j]} pattern). 
However, in \textsc{Operator 1}, the first stage breaks this limitation. See the pattern underscored and italicized in \cref{fig:kernel1-dfg}, which comprises 2 \prim{Share}s, 1 \prim{Reduce}, and 3 coordinates with domain $k_1$. The \prim{Share} in the first stage would have been \prim{Reduce}d, had it been a traditional convolution. But the \prim{Unfold}ed window remains and is passed to the second stage to be contracted with the weight. 

\begin{figure}
    \centering
    \includegraphics[width=\columnwidth]{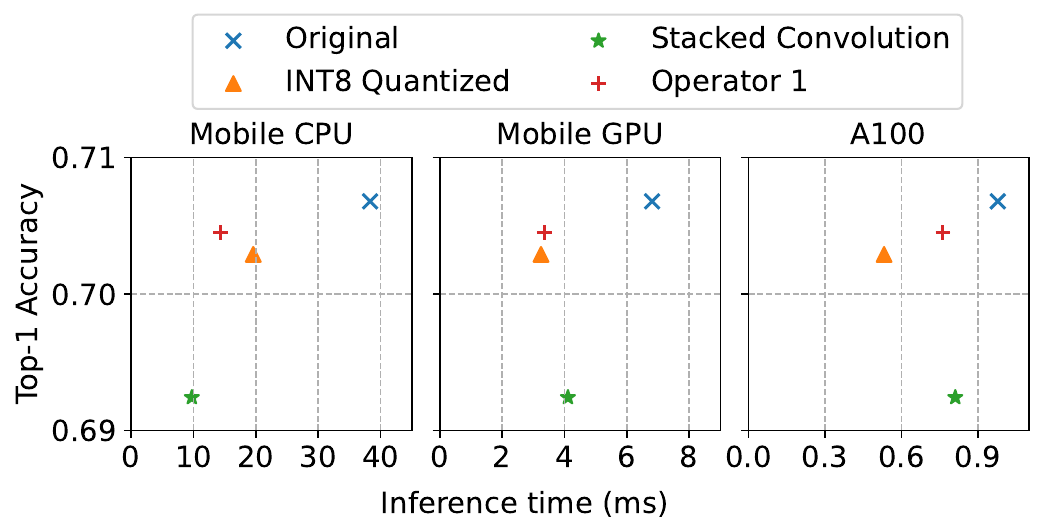}
    \caption{Comparison between \textsc{Operator 1} and other optimizations, evaluated on ImageNet with TVM.}
    \label{fig:kernel1_advantage}
\end{figure}

To see why such a pattern is effective, we stack two grouped convolutions into an operator, which is just \textsc{Operator 1} with the \prim{Share}d \texttt{k\_1} in stage 1 \prim{Reduce}d and the \texttt{W} in stage 2 \prim{Unfold}ed again (hence might be discoverable under traditional NAS schemes). As in \cref{fig:kernel1_advantage}, although having the same FLOPs and similar latency, the stacked convolution doubles the accuracy degradation, which we attribute to the difference in the receptive field in \textsc{Operator 1} ($3 \times 3$ vs. $3 \times 5$) that eases training of the model. This may provide insights for the machine learning community.

Since our design objective of trading accuracy for latency is the same as quantization, we further compare \sys with INT8 quantization. We obtain the quantized model from \texttt{torchvision.models.quantization.resnet18} using the QNNPACK configuration, and import it to TVM for inference optimizations.
As in \cref{fig:kernel1_advantage}, \textsc{Operator 1} has slightly better accuracy than INT8 quantization, as well as lower latency on the CPU. 
Note that \sys-synthesized operators can also be quantized, so the two techniques can be applied jointly to further enhance performance. Likewise, \sys can be combined with other similar techniques to achieve potentially more speedups.

\textsc{Operator 2} is a variant of \textsc{Operator 1}, and resembles two 1D convolutions with weights connected using \prim{Share} in a similar manner. 
Benefiting from the weight \prim{Share}-ing, \textsc{Operator 2} yields $6.19\times$, $3.27\times$, and $2.61\times$ speedups on the three hardware platforms, within 1.5\% accuracy loss on CIFAR-100. 
We attribute the substantial performance speedups to its fewer parameters (less than $1/4$ of standard 2D convolution) that can fit in the limited caches on edge devices.

\textbf{Common patterns.}
We identify several common patterns from the novel operators discovered by \sys. 
Aside from the convolution and grouping patterns visible in \textsc{Operator 1}, another common pattern is two weight tensors \prim{Share}-ing one or more dimensions, similar to low-rank decomposition, which is highly effective in reducing the number of parameters and FLOPs.
Also, we find multiple operators replacing one \prim{Unfold} on a spatial dimension with a \prim{Shift}, which can substantially reduce computations while still providing some extent of information mixture along this spatial dimension, similar to the idea in ShiftNet~\cite{wu2018shift}.
More patterns are unique to individual operators.

\begin{figure*}
    \centering
    \includegraphics[width=\textwidth]{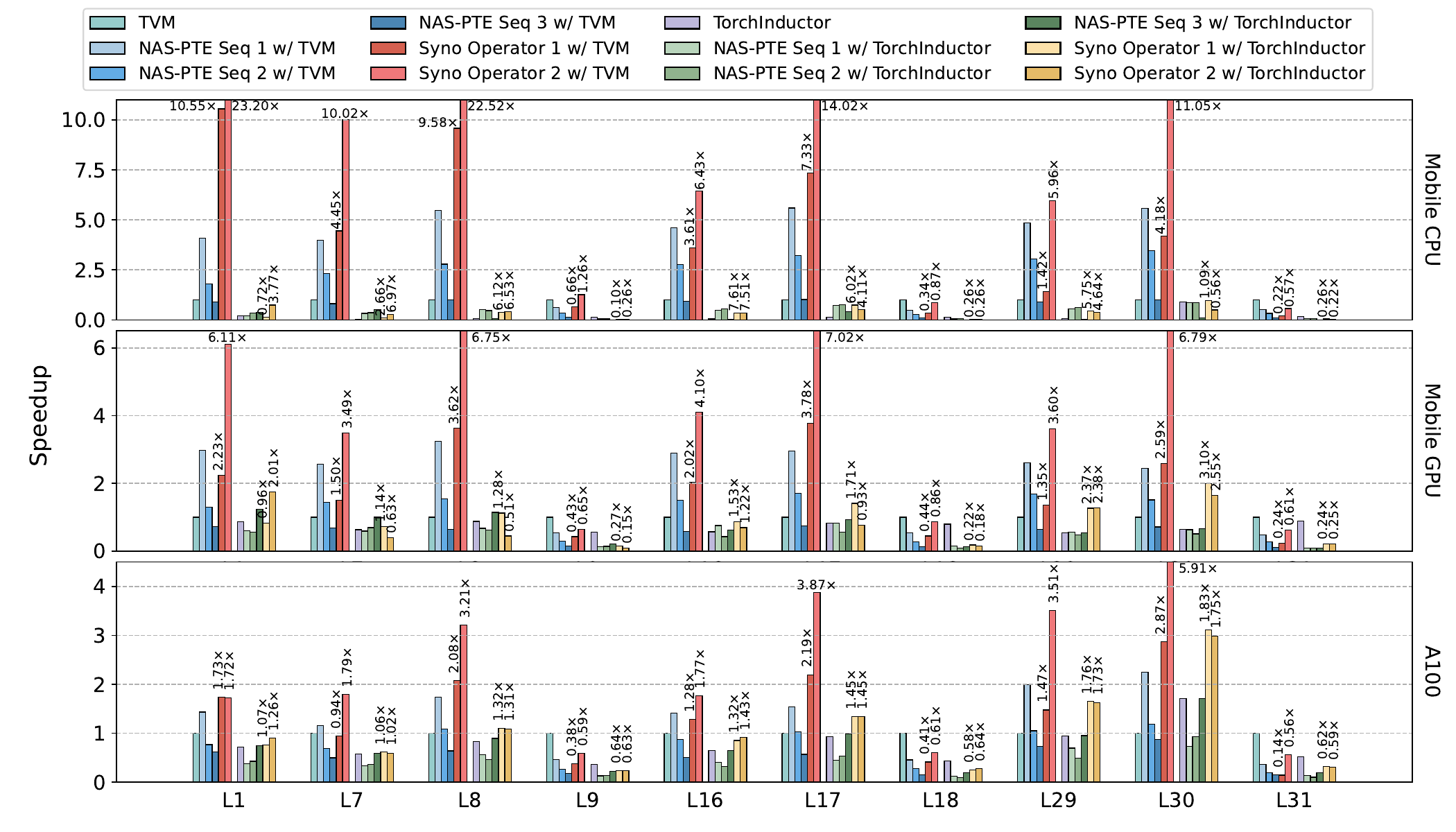}
    \caption{Layer-wise performance comparison between \sys and NAS-PTE on ResNet-34. The bars of each model are normalized to TVM.}
    \label{fig:layerwise}
\end{figure*}

\textbf{Comparison with NAS-PTE.}
\cref{fig:layerwise} shows the layer-wise performance of \textsc{Operator 1} and \textsc{Operator 2} compared to the original convolution and all the three operators proposed by NAS-PTE, when used in ResNet-34. 
On the mobile CPU, the mobile GPU, and A100, compared to NAS-PTE, the speedups of our best operators over their best ones are $2.13\times$, $1.68\times$, and $1.63\times$ on average when both are compiled with TVM, and $0.83\times$, $0.84\times$, and $1.38\times$ when both are compiled with TorchInductor.
Our best operators reduce the FLOPs by $1.76\times$ to $4.32\times$, and reduce the number of parameters by $1.80\times$ to $9.50\times$. 
The improvements are achieved without any layer-wise tuning like NAS-PTE but by a fully automated workflow in \sys.

Note that when compiled with TorchInductor, \sys underperforms NAS-PTE on the mobile CPU and GPU, despite the reduction on FLOPs and number of parameters. 
We find that TorchInductor often falls back to ATen kernels instead of generating native code for mobile hardware, as opposed to on A100 where it can emit efficient Triton code~\cite{tillet2019triton} in most of the time. The pre-compiled ATen kernels are less suitable for the novel \sys-generated operators. 
Actually, TorchInductor has mainly been optimized for large GPUs. Most of its templates target large GPUs, while smaller GPUs are neglected to keep the number of templates small and the compilation fast~\cite{isbiggpu}. 
Thus, we attribute the slowdown to the immaturity of TorchInductor on mobile CPUs and GPUs rather than the inability of \sys. The more generic compiler TVM is able to deliver consistent speedups.

\textbf{Comparison with $\mathbf{\alpha}$NAS.}
$\alpha$NAS reported FLOPs reduction ratios and training speedups for some variants of ResNet and EfficientNet. 
Within 2\% ImageNet accuracy drop, $\alpha$NAS achieves 25\% fewer FLOPs and about 12\% TPU-v3 training speedup on both ResNet-50 and EfficientNet-B0, while \sys achieves 63\% and 37\% fewer FLOPs and 56\% (48\%) and 12\% (7\%) A100 inference speedup when compiled with TVM (TorchInductor) on ResNet-34 and EfficientNet-V2-S, respectively. 
This qualitatively shows \sys's advantages. 

\begin{figure}
    \centering
    \includegraphics[width=\columnwidth]{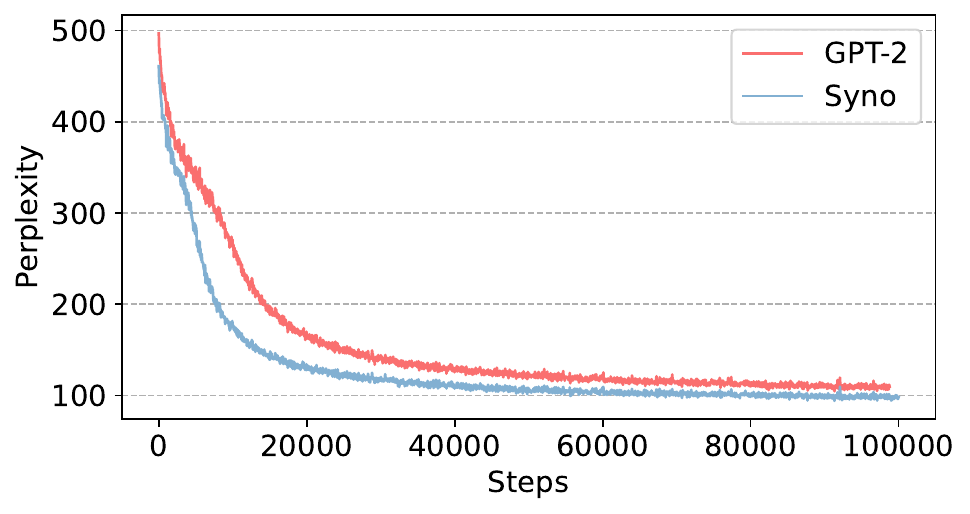}
    \caption{Comparison of language perplexity vs. training steps between \sys and the original GPT-2.}
    \label{fig:gpt-loss-result}
\end{figure}

\subsection{Results on GPT-2}

We follow Primer~\cite{primer} to allocate a 30-minute training period on GPT-2 for each searched operator and compare their final language perplexity results. 
We then extend the training for the best-performing operator and the original model to reach 100,000 steps as shown in \cref{fig:gpt-loss-result}.
When searching for substitutions for the QKV projections, our best operator achieves a $1.1\times$ training speedup and reduces the perplexity to $99$, outperforming the original model’s perplexity of $111$.
More specifically, our operator constructs the original projections by groups, which allows the QKV matrices used in the attention modules to learn from different features of input tokens, thereby improving the training efficiency.

\subsection{Ablation Studies}

\textbf{Canonicalization.}
To show the effectiveness of \sys canonicalization rules, we draw 6452 samples without canonicalization, among which only 86 are canonical. 
This implies that canonicalization cuts more than $70\times$ redundancy. 
\cref{tab:canonical-rate-table} shows the canonical rates for different pGraph sizes.

\begin{table}
\centering\tablefontsize
\caption{Canonical rates of different sampled pGraph sizes.}
\label{tab:canonical-rate-table}
\begin{tabular}{cccccccc}
\toprule
2      & 3      & 4      & 5      & 6      & 7      & $\ge$8 \\
\midrule
100.00\% & 18.18\% & 13.97\% & 4.40\% & 1.22\% & 0.08\% & 0.00\% \\
\bottomrule
\end{tabular}
\end{table}

\textbf{Shape distance.}
To verify the effectiveness of the shape distance metric, we evaluate the successful rates of random sample trials with and without the guidance of shape distance, respectively. 
On a server machine with 192 virtual cores, 253 distinct operators are found after 5 million trials in 68.33 seconds, with shape distance enabled. 
However, without using shape distance, 500 million trials in 180.51 seconds yield no valid operators. 
Thus, shape distance is vital for avoiding useless synthesis.

\section{Conclusions}

This paper advocates the paradigm of neural operator synthesis, which automatically discovers novel NN operators with good inference accuracy and/or execution speed. 
A practical framework named \sys has been implemented, using a rich set of fine-grained primitives to construct operators, applying canonicalization to eliminate redundancy, and guided by a novel operator shape distance metric to improve synthesis efficiency.
\sys is able to discover better NN operators than existing ones on various models, with higher execution performance and minor accuracy loss.

\begin{acks}
The authors thank the anonymous reviewers and our shepherd, Shoaib Kamil, for their valuable suggestions, and the Tsinghua IDEAL group members for constructive discussion.
Mingyu Gao is the corresponding author.
\end{acks}

\bibliographystyle{ACM-Reference-Format}
\bibliography{references}

\appendix
\section{Artifact Appendix}

\subsection{Abstract}

This artifact appendix helps the readers reproduce the main evaluation results of this paper. The artifact evaluation includes the instructions on how to synthesize operators with Syno, obtain the accuracy and performance results of the models with the novel operators, and plot Figures 5, 6, 8, 9, and 10. Our GitHub repository also provides a README containing detailed instructions.

\subsection{Artifact check-list (meta-information)}

{\small
\begin{itemize}
  \item {\bf Algorithm: }Syno's operator synthesis algorithm.
  \item {\bf Compilation: }GCC 13 and CUDA Toolkit 12.9. 
  \item {\bf Models: }ResNet-18, ResNet-34, DenseNet-121, ResNeXt-29-2x64D, EfficientNet-V2-S, and GPT-2. 
  \item {\bf Data sets: }CIFAR100, ImageNet, and lm1b.
  \item {\bf Run-time environment: }Linux Ubuntu 24.04. We provide a Dockerfile for the environment setup. 
  \item {\bf Hardware: }NVIDIA Jetson Orin Nano 8 GB board, and NVIDIA A100 GPUs.
  \item {\bf Metrics: }Model inference accuracy and execution latency.
  \item {\bf Output: }The key results are a list of synthesized operators discovered by Syno, plus five plots summarizing their performance. 
  \item {\bf Experiments: }The workflow is introduced in AE/README. 
  \item {\bf How much disk space required (approximately)?: }500GB (For ImageNet).
  \item {\bf How much time is needed to prepare workflow (approximately)?: }Docker images can be built within 30 minutes.
  \item {\bf How much time is needed to complete experiments (approximately)?: }Searching for operators requires around 300 GPU hours per model.
  Accuracy re-evaluation takes about 900 GPU hours for all models.
  Tuning for performance on the A100 and Jetson Orin Nano GPUs takes about 700 GPU hours each, and tuning on the Jetson Orin Nano CPU takes 700 hours.
  Plotting can be done within several minutes. 
  \item {\bf Publicly available?: }Yes.
  \item {\bf Code licenses (if publicly available)?: }MIT.
\end{itemize}
}

\subsection{Description}

\subsubsection{How to access}

\begin{itemize}
    \item Source code: \url{https://github.com/tsinghua-ideal/Syno}.
    \item Artifact evaluation data: \url{https://github.com/Yongqi-Zhuo/Syno-AE}, which is a git submodule of the above repository, so you need not separately clone it.
\end{itemize}

\subsubsection{Hardware dependencies}

\begin{itemize}
    \item Searching for operators requires A100 GPUs (an 8$\times$A100 machine is recommended).
    \item Performance tuning requires at least an A100 GPU and an NVIDIA Jetson Orin Nano 8 GB board.
\end{itemize}

\subsubsection{Software dependencies}

We provide a Dockerfile in our repository for easy reproduction, and the detailed software dependencies are listed there.

\subsubsection{Data sets}

The search and evaluation require three datasets: CIFAR100, ImageNet, and lm1b. CIFAR100 and lm1b will be automatically downloaded from TorchVision and Huggingface when executing our scripts. ImageNet requires some manual preparation, for which the detailed steps can be found in the README. 

\subsubsection{Models}

Our experiments are conducted with six models: ResNet-18, ResNet-34, DenseNet-121, ResNeXt-29-2x64D, EfficientNet-V2-S, and GPT-2. Note that you do not need the pre-trained weights for those models. 

\subsection{Installation}

Clone the repository and build the Docker image using the Dockerfile inside the repository.

\begin{verbatim}
  git clone --recursive \
    https://github.com/tsinghua-ideal/Syno.git
  docker build -t syno Syno
\end{verbatim}

The experiments will need preprocessed ImageNet. Download the dataset, format it into a PyTorch-style dataset, follow the instructions in \href{https://github.com/libffcv/ffcv-imagenet}{FFCV-ImageNet} to prepare the dataset with \verb|bash write_imagenet.sh 400 0.10 90|, and finally set the directory in Syno with \verb|bash set_imagenet_dir.sh |\\\verb|$WRITE_DIR|. 

\subsection{Experiment workflow}

On a high level, our experiments can be decoupled into four steps: 
\begin{enumerate}
    \item Searching: search for efficient operators to be substituted into the neural network models using the Syno operator synthesis algorithm. 
    \item Accuracy Evaluation: evaluate the accuracy of the models optimized by Syno on ImageNet. 
    \item Tuning: tune the Syno-optimized models with tensor compilers to obtain the performance numbers. 
    \item Plotting: plot the accuracy and performance results of the optimized models to visualize the tradeoff achieved with Syno.  
\end{enumerate}

Since these steps can take very long time, to facilitate easier reproduction, we provide our data in \verb|AE/data| as drop-in replacements for the results of each step, so the reproduction can start from any intermediate step.

\subsubsection{Searching}

Please use \verb|search.sh| for the search. 
Specifically, run \verb|bash search.sh $MODEL|, where \verb|$MODEL| is the model to search with. 
The supported models include 
\begin{itemize}
    \item \verb|resnet18|
    \item \verb|resnet34|
    \item \verb|resnext29_2x64d|
    \item \verb|densenet121|
    \item \verb|efficientnet_v2_s|
    \item \verb|gpt2|
\end{itemize}

\subsubsection{Accuracy Evaluation}

The search on the vision models will produce a list of operators with their CIFAR-100 accuracies and FLOPs, saved in \verb|AE/results/$MODEL-session|. 
After picking operators with good accuracies, you need to re-evaluate them on ImageNet by \verb|reevaluate_vision.sh|. 
The detailed instructions can be found in \verb|AE/README.md|. 

The search on GPT-2 will also produce a list of operators with their perplexity results after training for 30 minutes. After picking the best operator, re-evaluate it with 100,000 steps using \verb|reevaluate_gpt.sh|. 

Finally, we provide two scripts --- \verb|train_baseline.sh| and \verb|train_custom.sh|, with which you can obtain the accuracies for the baselines and the operators we picked for the case studies. 

\subsubsection{Tuning}

You need to set up the host, one or more A100 GPUs, and one or more NVIDIA Jetson Orin Nano's. Make sure the devices can access the host via internet connection. Also set up the TVM RPC trackers and servers according to \verb|AE/README.md|. Then run the grid tuners.
Refer to \verb|AE/README.md| for detailed instructions.

\begin{verbatim}
  # On host
  python grid_tune.py \
    --config /workspace/Syno/AE/grid_tune.json \
    --rpc-host $TRACKER_HOST \
    --rpc-port $TRACKER_PORT
  # On A100
  python grid_torch.py \
    --config /workspace/Syno/AE/grid_tune.a100.json
  # On NVIDIA Jetson Orin Nano
  python grid_torch.py \
    --config /workspace/Syno/AE/grid_tune.mdev.json
\end{verbatim}

\subsubsection{Plotting}

After the above steps, you can plot the figures with the experiment results. 

First, copy the tuning results: 
\begin{verbatim}
  bash copy_perf.sh mdev
  bash copy_perf.sh a100
\end{verbatim}
Then run \verb|bash plot.sh| to produce the figures. 

The script will produce 5 figures in \verb|AE/plots|:
\begin{enumerate}
    \item \verb|end-to-end-performance.pdf|: Figure 5.
    \item \verb|imagenet-performance.pdf|: Figure 6.
    \item \verb|case-study.pdf|: Figure 8.
    \item \verb|kernel-performance.pdf|: Figure 9.
    \item \verb|gpt-loss.pdf|: Figure 10.
\end{enumerate}

\subsection{Evaluation and expected results}

We provide our experiment results in \verb|AE/data|. If you use our data, then you should see exactly the same figures as in our paper. Otherwise, the numbers might be slightly different due to the randomness introduced by operator searching and the fluctuations of performance during tuning, but the overall trend should be the same.  

\subsection{Experiment customization}

You can write your own configuration files other than the provided \verb|AE/grid_tune.json| for performance tuning to use other hardware and other synthesized operators. See \verb|AE/README.md| for more details.

\subsection{Notes}

\subsection{Methodology}

Submission, reviewing and badging methodology:

\begin{itemize}
  \item \url{https://www.acm.org/publications/policies/artifact-review-and-badging-current}
  \item \url{https://cTuning.org/ae}
\end{itemize}

\end{document}